\documentclass[journal]{IEEEtran}
\usepackage{amsmath,amsfonts}
\usepackage{algorithmic}
\usepackage{algorithm}
\usepackage{array}
\usepackage[caption=false,font=normalsize,labelfont=sf,textfont=sf]{subfig}
\usepackage{textcomp}
\usepackage{stfloats}
\usepackage{url}
\usepackage{verbatim}
\usepackage{graphicx}
\usepackage{cite}
\usepackage{multirow}
\usepackage{tablefootnote}
\usepackage{multicol} 
\usepackage{arydshln}
\usepackage{xcolor}
\usepackage{times}
\usepackage{soul}
\usepackage{url}
\usepackage{makecell}
\usepackage[utf8]{inputenc}
\usepackage[small]{caption}
\usepackage[hidelinks]{hyperref}
\usepackage{graphicx}
\begin{document}

\title{Constraints Matrix Diffusion based Generative Neural Solver for Vehicle Routing Problems}

\author{Zhenwei Wang, Tiehua Zhang~\IEEEmembership{Member,~IEEE}, Ning Xue~\IEEEmembership{ Member,~IEEE}, Ender \"Ozcan~\IEEEmembership{Senior Member,~IEEE}, Ling Wang~\IEEEmembership{ Member,~IEEE}, Ruibin Bai~\IEEEmembership{Senior Member,~IEEE}\\

\thanks{Zhenwei Wang, Ning Xue and Ruibin Bai (Corresponding author) are with the School of Computer Science, University of Nottingham Ningbo China, Ningbo, 315100, China (email: zhenwei.wang@nottingham.edu.cn;
ning.xue@nottingham.edu.cn;ruibin.bai@nottingham.edu.cn).}
\thanks{Tiehua Zhang is with the School of Computer Science and Technology, Tongji University, Shanghai, 200092, China (email:tiehuaz@tongji.edu.cn).}
\thanks{Ender \"Ozcan is with the Computational Optimisation and Learning Lab,
School of Computer Science, University of Nottingham, NG8 1BB Nottingham,
U.K. (email:ender.ozcan@nottingham.ac.uk).}
\thanks{Ling Wang is with the Department of Automation,
Tsinghua University, Beijing 100084, China. (email:wangling@mail.tsinghua.edu.cn). }
}


\markboth{Journal of \LaTeX\ Class Files,~Vol.~14, No.~8, August~2025}%
{Shell \MakeLowercase{\textit{et al.}}: A Sample Article Using IEEEtran.cls for IEEE Journals}
\maketitle
\begin{abstract}
Over the past decade, neural network solvers powered by generative artificial intelligence have garnered significant attention in the domain of vehicle routing problems (VRPs), owing to their exceptional computational efficiency and superior reasoning capabilities. In particular, autoregressive solvers integrated with reinforcement learning have emerged as a prominent trend. However, much of the existing work emphasizes large-scale generalization of neural approaches while neglecting the limited robustness of attention-based methods across heterogeneous distributions of problem parameters. Their improvements over heuristic search remain largely restricted to hand-curated, fixed-distribution benchmarks. Furthermore, these architectures tend to degrade significantly when node representations are highly similar or when tasks involve long decision horizons. To address the aforementioned limitations, we propose a novel fusion neural network framework that employs a discrete noise graph diffusion model to learn the underlying constraints of vehicle routing problems and generate a constraint assignment matrix. This matrix is subsequently integrated adaptively into the feature representation learning and decision process of the autoregressive solver, serving as a graph structure mask that facilitates the formation of solutions characterized by both global vision and local feature integration. To the best of our knowledge, this work represents the first comprehensive experimental investigation of neural network model solvers across a 378-combinatorial space spanning four distinct dimensions within the CVRPlib public dataset. Extensive experimental evaluations demonstrate that our proposed fusion model effectively captures and leverages problem constraints, achieving state-of-the-art performance across multiple benchmark datasets.
\end{abstract}


\section{Introduction}
\IEEEPARstart{T}{he} rapid advancement of artificial intelligence (AI) over the past decade has fundamentally transformed the landscape of combinatorial optimization, particularly through the integration of generative AI techniques in vehicle routing problems (VRPs), thereby opening unprecedented opportunities for logistics, transportation, and operations optimization~\cite{li2022overview,bai2023analytics,bogyrbayeva2024machine,adache2024}. Exact methods (e.g., branch and bound, branch-and-price, cutting plane) provide optimality guarantees but, owing to the NP-hard nature of VRPs, become computationally intractable for obtaining optimal or near-optimal solutions at scale within practical time limits~\cite{xue2021hybrid}. Heuristic approaches construct initial solutions and iteratively refine them via neighborhood search~\cite{adache2023} and genetic operators, achieving strong performance under tight runtimes~\cite{zhang2022deep,zheng2021combining,yangBranchpriceandcutAlgorithmVehicle2021,zhaoHybridGeneticSearch2024}; however, they generalize poorly to uncertainties and typically require full re-optimization. By contrast, generative AI–driven neural solvers function as learned approximators that, once trained, deliver high-quality solutions on i.i.d. instances with fast online inference, thereby mitigating key limitations of conventional heuristics.

Autoregressive neural solvers, inspired by natural language processing methodologies, have undergone substantial evolution from their inception with Pointer Networks (PtrNets)~\cite{vinyalsPointerNetworks2017}, building upon the foundational framework of Recurrent Neural Networks (RNNs) with Long Short Term Memory (LSTM)~\cite{zarembaRecurrentNeuralNetwork2015}, to sophisticated encoder-decoder architectures leveraging Transformer or Graph Neural Network (GNN) based models. This line of research~\cite{vinyalsPointerNetworks2017,nazariReinforcementLearningSolving2018,koolAttentionLearnSolve2019,kwonPOMOPolicyOptimization2020,leiSolveRoutingProblems2022,wang2024gase,lingwang2024,zhao2025} has demonstrated remarkable convergence properties toward optimal solutions on synthetically generated datasets. Nevertheless, our experiments reveal a consistent pattern: autoregressive architectures that construct solutions from scratch have limited discriminative power when nodes share similar representations, largely because constraint features are not incorporated during representation learning. This limitation is exacerbated by the predominance of fully connected, global‑attention architectures~\cite{vaswaniAttentionAllYou2023}. Under uniform demands or strong spatial clustering, multi‑layer attention induces oversmoothing in node embeddings; conditioned on a context vector and these oversmoothed embeddings, the decoder produces dissolved attention logits, often degenerating toward near‑random trajectory sampling.

On the other hand, Reinforcement learning (RL) is appealing when (near-) optimal labels are impractical, but cannot reliably assess per‑step node decisions and instead updates policies from full trajectory rollouts. Under sparse, delayed rewards, errors accumulate with horizon length, limiting scalability to large, long‑horizon routing, motivating diffusion‑based approaches~\cite{song2020denoising,sun2023difusco,li2023t2t,li2024fast}. In these pipelines, high‑quality solutions are first generated by heuristics such as Hybrid Genetic Search (HGS)~\cite{vidal2022hybrid}. The diffusion model is trained by corrupting solutions with noise and learning denoising dynamics to predict posterior edge-coverage probabilities for unseen instances. Unlike autoregressive solvers, diffusion allows single-pass, length-independent inference, supporting efficient heuristic decoding and scalability. However, it still depends on high-quality labels and extensive post-processing (often complex search) to form valid tours, limiting its success to simple-constraint problems such as the Traveling Salesman Problem (TSP) and reducing performance on more complex variants.

Based on a comprehensive review of learning-based solvers, we identify weak constraint awareness as the primary source of the observed optimality gap. Most methods behave as greedy, likelihood‑maximizing selectors, i.e., a paradigm effective in Natural Language Processing (NLP)/Computer Vision (CV), where synonymy or small pixel perturbations rarely alter meaning or perception, but brittle in VRP, where even near‑identical single‑step choices can compound into large sequence errors around the boundaries of the constraint polyhedron. Moreover, decision quality hinges on learned node embeddings while the decoding procedure fails to explicitly enforce core feasibility constraints, creating a structural bottleneck that limits solution quality.

To address these limitations, we propose a constraint-mask guided fusion framework driven by graph-diffusion guidance. We first apply data augmentation to induce a many‑to‑one mapping from near‑optimal solutions across demand and geometric variations, from which one-hot constraint matrices are derived. Edge coverage in the matrices is then corrupted via a discrete noise transition process with a linear schedule, and a graph diffusion model is trained to denoise and reconstruct the original constraint matrices, aiming for a fast inference of constraint pattern from experiences. The reconstructed matrices serve as prior topological masks for an autoregressive encoder–decoder. The predicted constraint masks gate local representation learning in the encoder and are fused with global representations, thereby mitigating oversmoothing in fully connected graph structures while preserving global context for decision making.
The contributions of this work can be summarized as follows:
\begin{itemize}
    \item To the best of our knowledge, this is the first comprehensive, systematic evaluation of autoregressive neural solvers across all 378 category combinations spanning four dimensions in the XML100 dataset from the CVRPLIB repository~\cite{queiroga2022}, thereby revealing intrinsic limitations of contemporary neural solvers for VRP.
    \item We propose a novel graph diffusion guided constraint‑matrix masking scheme that integrates local and global representations to mitigate oversmoothing in fully connected graph encoders. Constraint matrices, reconstructed from (near-) optimal solutions via graph diffusion, serve as topological masks that strengthen local representation learning and decision‑making; the resulting local embeddings are fused with global features, enhancing the autoregressive model’s performance and robustness.
    \item Extensive experiments and analyses demonstrate that our graph diffusion prior mask fusion framework attains state-of-the-art performance on both synthetic datasets and multi-dimensional, cross-distribution benchmarks that closely reflect real-world conditions.
\end{itemize}
\section{Related Work}
In this section, we review generative AI approaches to the VRP, focusing on three paradigms: autoregressive generation, diffusion-based generation, and improvement heuristics.
\subsection{Auto-Regressive Neural Solvers for VRPs}
Inspired by variable-length sequence mapping in NLP, autoregressive VRP models have advanced substantially. Pointer Networks established the seminal encoder–decoder paradigm for supervised TSP~\cite{vinyalsPointerNetworks2017}. Nazari et al. extended this line with LSTMs and introduced the first RL-based unsupervised VRP solver~\cite{nazariReinforcementLearningSolving2018}. Kool et al. proposed a Transformer-based, position-free encoder with REINFORCE self-critical training, achieving major gains~\cite{koolAttentionLearnSolve2019}. Building on Transformers, POMO~\cite{kwonPOMOPolicyOptimization2020} further improved performance via data augmentation and multi-start, parallel, average-baseline RL, becoming the backbone for many subsequent approaches. Follow-ups refined POMO along data scale and cross-problem generalization axes~\cite{gao2023towardslocalglobal,bi2022learning,luo2023neural,huang2025rethinking}. A parallel direction employs graph neural encoders, e.g., multi-layer Graph Attention Networks (GAT)~\cite{velickovicGraphAttentionNetworks2018a} with edge weights, followed by autoregressive decoding~\cite{wang2024gase,leiSolveRoutingProblems2022,fellekGraphTransformerReinforcement2023}; Wang et al. further leverage hypergraph learning to encode constraints via dynamically constructed hyperedges, enabling cross-problem solving~\cite{wang2025towards}.

Despite strong results, these approaches still struggle to internalize constraint structure: at inference, they largely perform greedy, likelihood-based decoding over attention-weighted node embeddings within a single distribution, limiting their ability to reason explicitly about constraints across diverse settings.
\subsection{Solving VRPs via Diffusion-Based Neural Approaches}
Another prominent line of work leverages diffusion models, which have achieved notable success in computer vision and, more recently, in routing. Sun et al.~\cite{sun2023difusco} apply discrete and continuous forward corruption to optimal TSP tours, learn timestep-wise noise mappings, and perform reverse denoising from random initialization to reconstruct solutions. Similarly, Li et al.~\cite{li2023t2t,li2024fast} generate initial TSP solutions via diffusion and refine them through post-processing. Conceptually, these approaches corrupt high-quality solutions and train denoisers to minimize a variational upper bound on negative log-likelihood, yielding edge-probability heatmaps that capture solution structure. Diffusion thus offers one-shot heatmap generation and strong scalability, with promising results on TSP. However, its applicability remains limited for constraint-rich problems. For example, in CVRP, capacity constraints fragment the solution space: instead of a single flow-balancing Hamiltonian cycle in TSP, the task becomes finding Hamiltonian paths within connected components induced by sparse graph cuts. This constraint-driven partitioning makes diffusion models prone to infeasible outputs, even with posterior search.

\subsection{Improvement Heuristic Neural Solvers for VRPs}
Improvement heuristics iteratively destroy and repair feasible solutions. A neural model typically first proposes a feasible solution, which is then enhanced via heuristic search; alternatively, the destroy/repair operators themselves are learned. For example, Joshi et al.~\cite{joshi2019efficient} use graph convolutional networks to learn node embeddings and derive heatmaps that guide iterative heuristic search. Gao et al.~\cite{gaoLearnDesignHeuristics2020} employ a GAT encoder to fuse node–edge features and a GRU-based decoder to generate destroy–repair operator pairs, training within an actor–critic framework. Deudon et al.~\cite{deudonLearningHeuristicsTSP2018} predict TSP tours via RL with a critic and subsequently refine them using operators such as 2-opt. Chen et al.~\cite{chen2019learning} train a policy to select rewrite operators that reconstruct solution sequences and progressively improve quality. While effective across diverse combinatorial problems, these methods do not directly predict optimal solutions; they depend on search-based destruction and reconstruction, incurring notable efficiency costs and limiting real-time applicability. Moreover, as with diffusion approaches heavily reliance on post-search in complex, constraint-rich settings can obscure the net contribution of the neural component sophisticated strategies such as Monte Carlo Tree Search (MCTS)~\cite{swiechowski2023monte} often yield strong results even when paired with simple heuristics.
\section{Preliminaries}
\subsection{Problem Statement}
Given a graph $\mathcal{G}=(\mathcal{V}, \mathcal{A})$, where $\mathcal{V}$ represents the node set, comprising a depot node and $N$ customer nodes to be served, and $\mathcal{A}$ denotes the edge set between all nodes. The Capacitated Vehicle Routing Problem (CVRP) requires that, collectively, a fleet of vehicles visit each node exactly once while ensuring that the cumulative demands from all nodes serviced by each vehicle do not exceed its capacity $Q$. The objective is to minimize the total distance of all vehicles. 
The mathematical formulation can be expressed as follows:

\begin{equation}\label{eq1}
    \min \sum_{r \in \Omega} c_r z_r
\end{equation}
s.t.
\begin{equation}\label{eq2}
    \sum_{r \in \Omega} z_{r}\le|\mathcal{K}|
\end{equation}
\begin{equation}\label{eq3}
    \sum_{r \in \Omega} a_{i,r} z_r=1 ,\quad \forall i \in \mathcal{V} \backslash\{0\}
\end{equation}

Formula (\ref{eq1}) defines the optimization objective of the CVRP, where $c_r$ represents the cost associated with subgraph Hamiltonian circuit $r$ covered by vehicles, and $z_r$ denotes a binary decision variable that determines whether $r$ is selected to form the current solution within the set of all feasible subgraph Hamiltonian circuits $\Omega$ that satisfy capacity constraints. Constraint (\ref{eq2}) imposes a limitation on the number of vehicles, stipulating that the cumulative sum of route $r$ formed by vehicles in the current solution must not exceed the total number of available vehicles $\mathcal{K}$. It is noteworthy that in contemporary autoregressive-based neural solvers, this constraint is often relaxed. Constraint (\ref{eq3}) ensures that all customer nodes, except node 0 (depot), are visited exactly once, where $a_{i,r}$ is a binary constant to indicate whether node $i$ is included in vehicle route $r$ or not. 

\subsection{Constraint Matrix}
In this work, we define a binary constraint matrix \( \mathcal{M} \in \{0,1\}^{N\times N} \) where \( \mathcal{M}_{ij}=1 \) if nodes \( i \) and \( j \) belong to the same vehicle tour (Hamiltonian circuit), and \( \mathcal{M}_{ij}=0 \) otherwise. The matrix satisfies:
(1) \textbf{Symmetry}: \( \mathcal{M}_{ij}=\mathcal{M}_{ji} \). (2) \textbf{Transitivity}: if \( \mathcal{M}_{ij}=1 \) and \( \mathcal{M}_{jk}=1 \), then \( \mathcal{M}_{ik}=1 \). (3) \textbf{No self-edges}: \( \mathcal{M}_{ii}=0 \). (4) \textbf{Capacity consistency}: for each node \( i \), the total demand of the set \( \{i\} \cup \{ j:\mathcal{M}_{ij}=1 \} \) does not exceed the vehicle capacity \( Q \).

Given an optimal or near-optimal solution, \( \mathcal{M} \) can be directly constructed from the corresponding route memberships. The diffusion model is then trained to predict such constraint matrices for previously unseen instances.
\subsection{Markov Decision Process}
After learning the constraint matrix, the diffusion model incorporates this information with graph embeddings as a topological mask, embedding the constraint structure into the encoder and mitigating oversmoothing of node representations. Decision-making is formalized as an MDP and trained autoregressively through RL.

\subsubsection{State}
We decompose the state into static and dynamic components. The static state remains fixed throughout a rollout and comprises node embeddings—learned representations of demands, coordinates, and edge weights. The dynamic state is updated after each action and includes fused embeddings of remaining vehicle capacity, visitation status, and other step-dependent variables.

\subsubsection{Action}
An action selects the next node to visit. During training, we employ sampling; at inference, a greedy policy is used. Concretely, the decoder computes attention scores between the current state embedding and node embeddings, which are normalized into a categorical distribution for sampling. Let \( \theta \) denote network parameters and \( \pi \) the policy; the probability of a trajectory \( \tau \) on instance \( \mathcal{G} \) factorizes as
\begin{equation}\label{eq5}
    p_{\pi_{\theta}}(\tau \mid \mathcal{G}) = \prod_{t=1}^{|\tau|} p_{\pi_{\theta}} \big(\tau_t \mid \mathcal{G}, \tau_{t^\prime}, \forall\, t^\prime < t \big).
\end{equation}
At inference, the action with the highest probability \( p_{\pi_{\theta}} \big(\tau_t \mid \mathcal{G}, \tau_{t^\prime}, \forall\, t^\prime < t \big) \) is selected at each step \( t \).

\subsubsection{Reward}
Upon the completion of routing process, the trajectory reward \( \mathcal{R} \) is computed, and policy gradients~\cite{suttonPolicyGradientMethods1999} are used to optimize \( \theta \). The standard reward is the negative tour length, so reward maximization is equivalent to distance minimization:
\begin{equation}\label{rewardcal}
    \mathcal{R}(\tau) = -L(\tau) = -\sum_{i=0}^{|\tau|}\left\|\tau_{i}-\tau_{i+1}\right\|_2.
\end{equation}
\section{Methodology}
This section presents our proposed fusion methodology from the following perspectives: Data Augmentation, graph diffusion-based constraint matrix generation, and topological mask-guided encoding and decoding. 
\subsection{Data Augmentation}
\begin{figure*}[htb]
    \centering
    \includegraphics[width=1.0\linewidth]{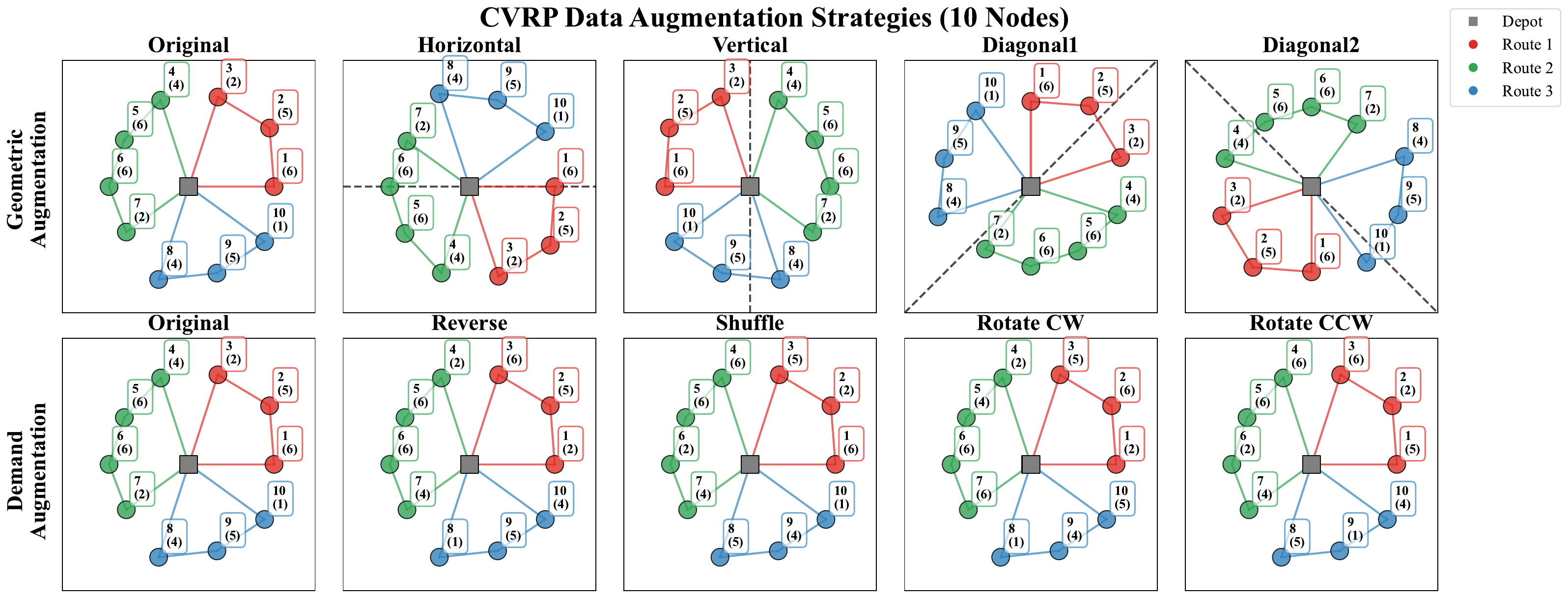}
    \caption{Demonstrating the effects of data augmentation, the numerical value within the rectangle above each scatter point indicates the origin node index, while the value in parentheses represents the node demand. The black dashed lines in the first row of the four symmetric transformation subgraphs correspond to the symmetry axes.}
    \label{fig:dataaug}
\end{figure*}
Before feeding each instance into the neural architecture, we apply two augmentation modalities: geometric augmentation of node coordinates and demand augmentation informed by network characteristics. We combine them to induce a many-to-one mapping from instances to optimal solutions, thereby improving robustness. The procedure is illustrated in Fig.~\ref{fig:dataaug}.

For geometric augmentation, inspired by Yao et al.~\cite{yaoRethinkingSupervisedLearning2024}, we use four axial symmetries: horizontal, vertical, and the two diagonals. These are implemented by centering the graph, converting to polar coordinates, and applying reflections/rotations. The first row of Fig.~\ref{fig:dataaug} shows that the optimal solution is invariant under these transformations. For demand augmentation, we exploit the observation that if demands within a subpath satisfy capacity constraints, permuting those demands within the subpath preserves optimality. Accordingly, we adopt four strategies in sub-routes: demand inversion, random reassignment, clockwise cyclic permutation (CW), and counter-clockwise cyclic permutation (CCW) (second row of Fig.~\ref{fig:dataaug}).

\subsection{Graph Diffusion Constraint Matrix Generation}
This subsection introduces the supervised graph diffusion framework, comprising constraint‑matrix generation, forward noise injection, and graph denoising.
\subsubsection{Constraint Matrices Generation}
\begin{figure}[hbt]
    \centering    \includegraphics[width=1.0\linewidth]{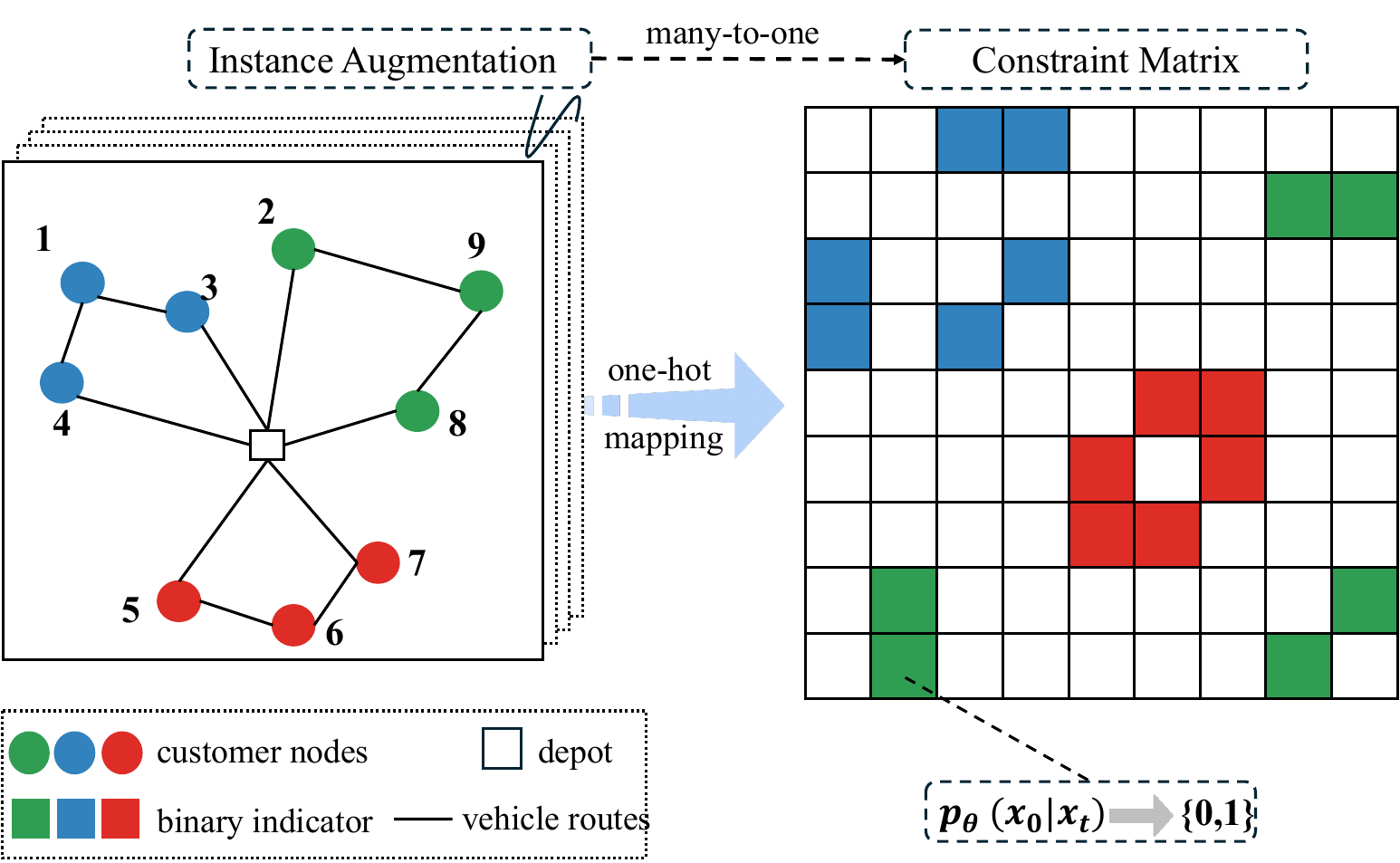}
    \caption{Constraint matrix mapping from augmented optimal solution instances}
    \label{fig:constraint_matrix}
\end{figure}

An optimal CVRP tour naturally decomposes into variable-length subpaths (routes). Because each customer is visited exactly once, we explicitly encode subpath membership with an $N\times N$ binary matrix, where entry $(i,j)=1$ if and only if customers $i$ and $j$ lie on the same routes. For example, for the visit sequence $0\!-\!4\!-\!1\!-\!3\!-0\!-\!2\!-\!9\!-\!8\!-0\!-\!5\!-\!6\!-7$ with depot $0$ and nine customers, a $9\times9$ sparse matrix captures the customer pairs that satisfy the capacity constraint under the optimal solution (Fig.~\ref{fig:constraint_matrix}). 
The generated constraint matrix serves as the training label for the graph diffusion model, essentially representing the original solution to be reconstructed. Under the aforementioned augmentations, instances with different coordinates and demands map to identical constraint matrices.

\subsubsection{Diffusion Process}
\begin{figure*}[htb]
    \centering
    \includegraphics[width=1.0\linewidth]{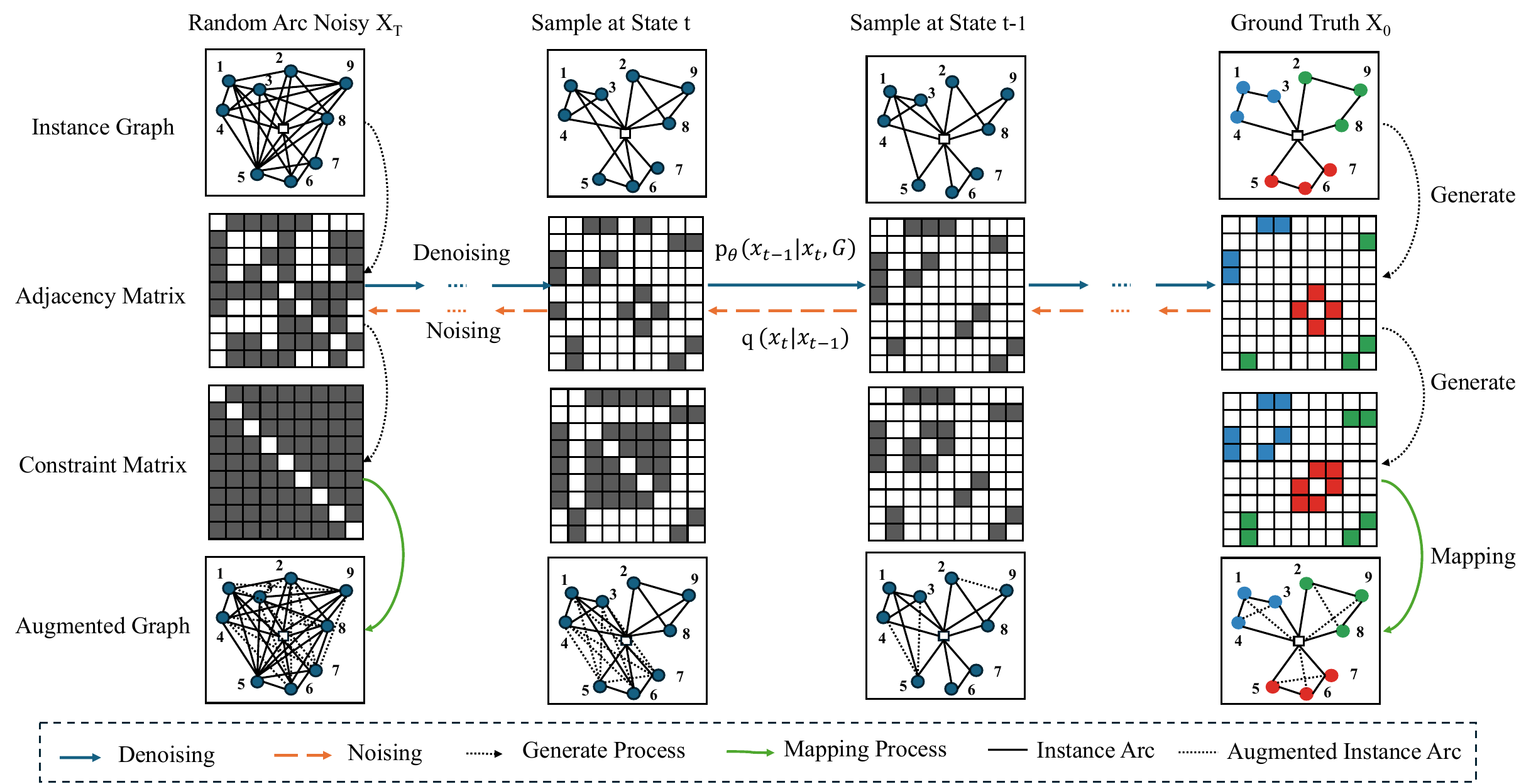}
    \caption{Constraint-Matrix Generation and Forward Corruption/Reverse Denoising under a Discrete-Noise Diffusion}
    \label{fig:diffusionprocess}
\end{figure*}

The diffusion model corrupts the original constraint matrices $\mathbf{x}_0\sim q(X_0\mid\mathcal{G})$ via a forward process, gradually producing pure noise $\mathbf{x}_T$. At different temporal steps $t$, the corresponding relationship between data and noise, specifically $\mathbf{x}_1...\mathbf{x}_T$ is essentially a latent variable with identical dimensionality to $\mathbf{x}_0$, implying that $q(\mathbf{x}_{t-1}|\mathbf{x}_0,\mathbf{x}_T)$ represents the known forward condition. Learning proceeds by approximating the reverse dynamics $p(\mathbf{x}_{t-1}\mid \mathbf{x}_t)$. With model parameters $\theta^\prime$, the marginal
$p_{\theta^\prime}(\mathbf{x}_0)=\int p_{\theta^\prime}(\mathbf{x}_{0:T})\,d\mathbf{x}_{1:T}$
leads to the training objective $\mathbb{E}\big[-\log p_{\theta^\prime}(\mathbf{x}0)\big]$, which measures how well $p_{\theta^\prime}(\cdot)$ fits $q(\mathbf{x}_0\mid\mathcal{G})$. Since direct computation of high-dimensional integrals is typically computationally intractable, the actual optimization objective is to approximate this using a variational lower bound, specifically:
\begin{equation}\label{variaenclb}
    \mathbb{E}\left[-\log p_{\boldsymbol{\theta^\prime}}\left(\mathbf{x}_0\right)\right]  \leq \mathbb{E}_q\left[-\log \frac{p_{\boldsymbol{\theta^\prime}}\left(\mathbf{x}_{0: T}\right)}{q\left(\mathbf{x}_{1: T} \mid \mathbf{x}_0\right)}\right]
\end{equation}
where $p_{\boldsymbol{\theta^\prime}}\left(\mathbf{x}_{0: T}\right)=p\left(\mathbf{x}_T\right) \prod_{t=1}^T p_{\boldsymbol{\theta^\prime}}\left(\mathbf{x}_{t-1} \mid \mathbf{x}_t\right)$ represents the denoising process joint probability distribution that gradually reconstructs the original data while distribution $q\left(\mathbf{x}_{1: T} \mid \mathbf{x}_0\right)=\prod_{t=1}^T q\left(\mathbf{x}_t \mid \mathbf{x}_{t-1}\right)$ can be interpreted as a known forward noise process. The variational lower bound can be expanded by the KL divergence between two distributions $p,q$, the conditional probability of reconstructing the original constraint matrix $\mathbf{x_0}$ from the first-step noise latent variable $\mathbf{x_1}$ in the denoising process, and a constant term $C$, providing an optimizable approximate objective for the negative log-likelihood that is computationally intractable to evaluate directly, i.e.:
\begin{equation}\label{dufloss}
\begin{aligned}
    \mathcal{L} =& \mathcal{C} + \mathbb{E}_q[-\log p_{\boldsymbol{\theta^\prime}}\left(\mathbf{x}_0 \mid \mathbf{x}_1\right )  \\
    +&\sum_{t>1} D_{K L}\left[q\left(\mathbf{x}_{t-1} \mid \mathbf{x}_t, \mathbf{x}_0\right) \| p_{\boldsymbol{\theta^\prime}}\left(\mathbf{x}_{t-1} \mid \mathbf{x}_t\right)\right]]
\end{aligned}
\end{equation}
Note that in equation (\ref{dufloss}), the KL divergence term quantifies the discrepancy between two probability distributions. The distribution $q(\mathbf{x}_{t-1}|\mathbf{x}_t,\mathbf{x}_0)$ represents the true conditional distribution derived from the forward noise, and $p_{\theta^\prime}(\mathbf{x}_{t-1}|\mathbf{x}_t)$ denotes the conditional probability distribution learned by the model during the denoising process. The summation term represents the cumulative discrepancy between the conditional distribution of the model's reverse process and the true conditional distribution derived from the forward process across all steps where $t > 1$. The term $-\log p_{\theta^\prime}(\mathbf{x_0}|\mathbf{x_1})$ quantifies the difficulty of the model directly reconstructing the original data from the first-step noise. The constant term $\mathcal{C}$ is independent to the model parameters $\theta^\prime$ and can be neglected during the optimization process, as it does not affect the parameter update direction. By minimizing this lower bound (i.e., reducing the sum of KL divergences and $-\log p_{\theta^\prime}(\mathbf{x_0}|\mathbf{x_1})$), the model can indirectly achieve better fitting to the data distribution, ultimately enabling the learned reverse denoising process to accurately reconstruct samples that closely approximate the true constraint matrices distribution from noise.  Fig.~\ref{fig:diffusionprocess} illustrates the forward corruption and reverse denoising procedures, conditioned on constraint-matrix labels derived from the underlying graph; the next two subsections detail noise injection and denoising.
\paragraph{Forward Noising}
When injecting noise into the original adjacency matrix, the constraint matrix $\mathbf{x}_0 \in \{0,1\}^{N \times N}$ is initially one-hot encoded to obtain $\mathbf{\bar{x}}_0 \in \{0,1\}^{N \times N \times 2}$. Discrete noise is subsequently added through the noise state transition matrix $\mathbf{Q}_t=\left[\begin{array}{cc}
\left(1-\beta_t\right) & \beta_t \\
\beta_t & \left(1-\beta_t\right)
\end{array}\right] \in \{0,1\}^{2 \times 2}$ according to the corruption parameter $\beta$ at each temporal step $t$, where $\beta$ follows a linear schedule with respect to $t$, i.e. $\beta_1 = 1e^{-4},\beta_T = 2e^{-2} $. Consequently, the one-step conditional distribution is $q(\mathbf{x}_t|\mathbf{x}_{t-1}) = \operatorname{Cat}(\mathbf{x}_t; \mathbf{p} = \bar{\mathbf{x}}_{t-1}\mathbf{Q}_t)$, where $\operatorname{Cat}(\cdot;\mathbf{p})$ denotes the categorical distribution with parameter $\mathbf{p}$ representing the probability vector. Following $T$ steps of noise injection, the cumulative $\mathbf{\bar{Q}}_T$ = $\mathbf{Q}_1\mathbf{Q}_2...\mathbf{Q}_t = \left[\begin{array}{cc}
0.5 & 0.5 \\
0.5 & 0.5
\end{array}\right]$. The T-step marginal distribution is expressed as: $q(\mathbf{x_T}|\mathbf{x_0}) = \operatorname{Cat}(\mathbf{x_T;p=\mathbf{\bar{x}_0}\mathbf{\bar{Q}}_T})$, leading to a state with maximum entropy, whereby the constraints in the original constraint matrix become disordered. At this stage, the probability transition is independent of the sparsity characteristics of $\mathbf{x}_0$ and demonstrates applicability across all items in $\mathbf{x}_0$. Furthermore, according to Bayes' theorem, given $\mathbf{x}_0$ and $\mathbf{x_t}$, the posterior distribution probability of $\mathbf{x}_{t-1}$ is expressed as:
\begin{equation}\label{forwardnoising}
\begin{aligned}
    q\left(\mathbf{x}_{t-1} \mid \mathbf{x}_t, \mathbf{x}_0\right)=&\frac{q\left(\mathbf{x}_t \mid \mathbf{x}_{t-1}, \mathbf{x}_0\right) q\left(\mathbf{x}_{t-1} \mid \mathbf{x}_0\right)}{q\left(\mathbf{x}_t \mid \mathbf{x}_0\right)}\\
    =&\operatorname{Cat}\left(\mathbf{x}_{t-1} ; \mathbf{p}=\frac{\bar{\mathbf{x}}_t \mathbf{Q}_t^{\top} \odot \bar{\mathbf{x}}_0 \bar{\mathbf{Q}}_{t-1}}{\bar{\mathbf{x}}_0 \bar{\mathbf{Q}}_t \bar{\mathbf{x}}_t^{\top}}\right)
\end{aligned}
\end{equation}
where the numerator $\bar{\mathbf{x}}_t \mathbf{Q}_t^{\top}$ is combined with $\bar{\mathbf{x}}_0 \bar{\mathbf{Q}}_{t-1}$ through element-wise multiplication $\odot$, corresponding to the parameters of the conditional probability $q\left(\mathbf{x}_t \mid \mathbf{x}_{t-1},\mathbf{x}_0\right)$ and the marginal distribution $q\left(\mathbf{x}_{t-1} \mid \mathbf{x}_0\right)$, respectively. The denominator $\bar{\mathbf{x}}_0 \bar{\mathbf{Q}}_t \bar{\mathbf{x}}_t^{\top}$ serves as a normalization factor that ensures the probabilities sum to unity, representing the probability value corresponding to $\mathbf{x}_t$ in $q\left(\mathbf{x}_t \mid \mathbf{x}_0\right)$.
\paragraph{Anisotropic Graph Denoising }
The denoising process employs a neural network to reconstruct $\mathbf{x}_0$ from the sampled $\mathbf{x}_t$. This constitutes the inverse of the forward noising process. Herein, we train an anisotropic graph neural network and utilize its gating mechanism, as described in~\cite {sun2023difusco}, to learn the mapping between $\mathbf{x}_0$ and the model's predicted $\hat{\mathbf{x}}_0$ through loss function (\ref{dufloss}). During inference, the predicted $\hat{\mathbf{x}}_0$ is substituted for the true $\mathbf{x}_0$ to complete the denoising process:
\begin{equation}
    p_{\boldsymbol{\theta}^\prime}\left(\mathbf{x}_{t-1} \mid \mathbf{x}_t\right)\propto\sum_{\hat{\mathbf{x}}_0} q\left(\mathbf{x}_{t-1} \mid \mathbf{x}_t, \hat{\mathbf{x}}_0\right) p_{\boldsymbol{\theta}^\prime}\left(\hat{\mathbf{x}}_0 \mid \mathbf{x}_t\right)
\end{equation}
The parameter $\boldsymbol{\theta^\prime}$ is derived from a multi-layer gated anisotropic GNN. The network input comprises the node representation embeddings learned by a pre-trained GAT, and the edge representation embeddings fused with temporal step encoding and $\mathbf{x}_t$. The output constitutes the probability distribution of the predicted constraint matrix. The node and edge representation are updated through Equations (\ref{edgel1})-(\ref{nodeupdate}):
\begin{equation}\label{edgel1}
    \boldsymbol{e}_{i j}^{\ell+1}=\boldsymbol{W_1}^{\ell} \boldsymbol{h}_i^{\ell}+\boldsymbol{W_2}^{\ell} \boldsymbol{h}_j^{\ell}+\boldsymbol{W_3}^{\ell} \boldsymbol{\bar{e}}_{i j}^{\ell}
\end{equation}
\begin{equation}\label{edgel}
\boldsymbol{\bar{e}}_{ij}^{\ell}=\operatorname{MLP}_e\left(\operatorname{BN}\left(\boldsymbol{e}_{i j}^{\ell}\right)\right)+\mathrm{MLP}_t(\mathbf{\bar{t}})
\end{equation}
\begin{equation}\label{gate}
    \mathbf{gate}_{ij}^{\ell+1} = \sigma(\mathbf{e}_{ij}^{\ell+1})\odot \boldsymbol{W_4}\boldsymbol{h}_j^{\ell}
\end{equation}
\begin{equation}\label{nodeupdate}
    \mathbf{h}_i^{\ell+1} = \mathbf{h}_i^{\ell} + \operatorname{ReLU}(\operatorname{BN}(\boldsymbol{W_5}^{\ell}\boldsymbol{h}_i^{\ell}+\sum_{j \in \mathcal{N}_{(i)}}\mathbf{gate}_{ij}^{\ell+1}))
\end{equation}
The learnable parameters integrate the edge features ($\bar{e}_{ij}^{\ell}$), temporal step positional encoding $\bar{\mathbf{t}}$, source node features ($h_i^{\ell}$), and target node features ($h_j^{\ell}$), thereby capturing anisotropy (distinguishing the influence of source/target nodes). $\mathbf{W_1}$ to $\mathbf{W_5}$ represent the learnable weight parameters, and $\operatorname{MLP}(\cdot)$ denotes a two-layer multilayer perception(MLP). Note that $h^0 = \operatorname{GAT}(\mathcal{G})$, which is extracted from a pretrained five-layer GAT, while $e^0 = {\mathbf{x}}_t$ is assigned by the noising process. $\sigma$ is the sigmoid function while $\operatorname{ReLU}(\cdot)$ and $\operatorname{BN}(\cdot)$ denote ReLU activation and Batch Normalization(BN), respectively. The temporal step positional encoding $\bar{\mathbf{t}}$ is defined as:
\begin{align}
\text{PE}{(t, 2i)} &= \sin\left( t \cdot 10000^{-2i / d} \right) \\
\text{PE}{(t, 2i+1)} &= \cos\left( t \cdot 10000^{-2i / d} \right)
\end{align}
Following training, the graph diffusion model ultimately predicts a constraint matrix, and the performance Area Under Curve (AUC) value on the test set exceeds 0.95, indicating robust discrimination of same-subpath node assignments in sparse constraint matrices.

\subsection{Topological Mask-Guided Encoder-Decoder}
\begin{figure*}[htb]
    \centering
    \includegraphics[width=0.99\linewidth]{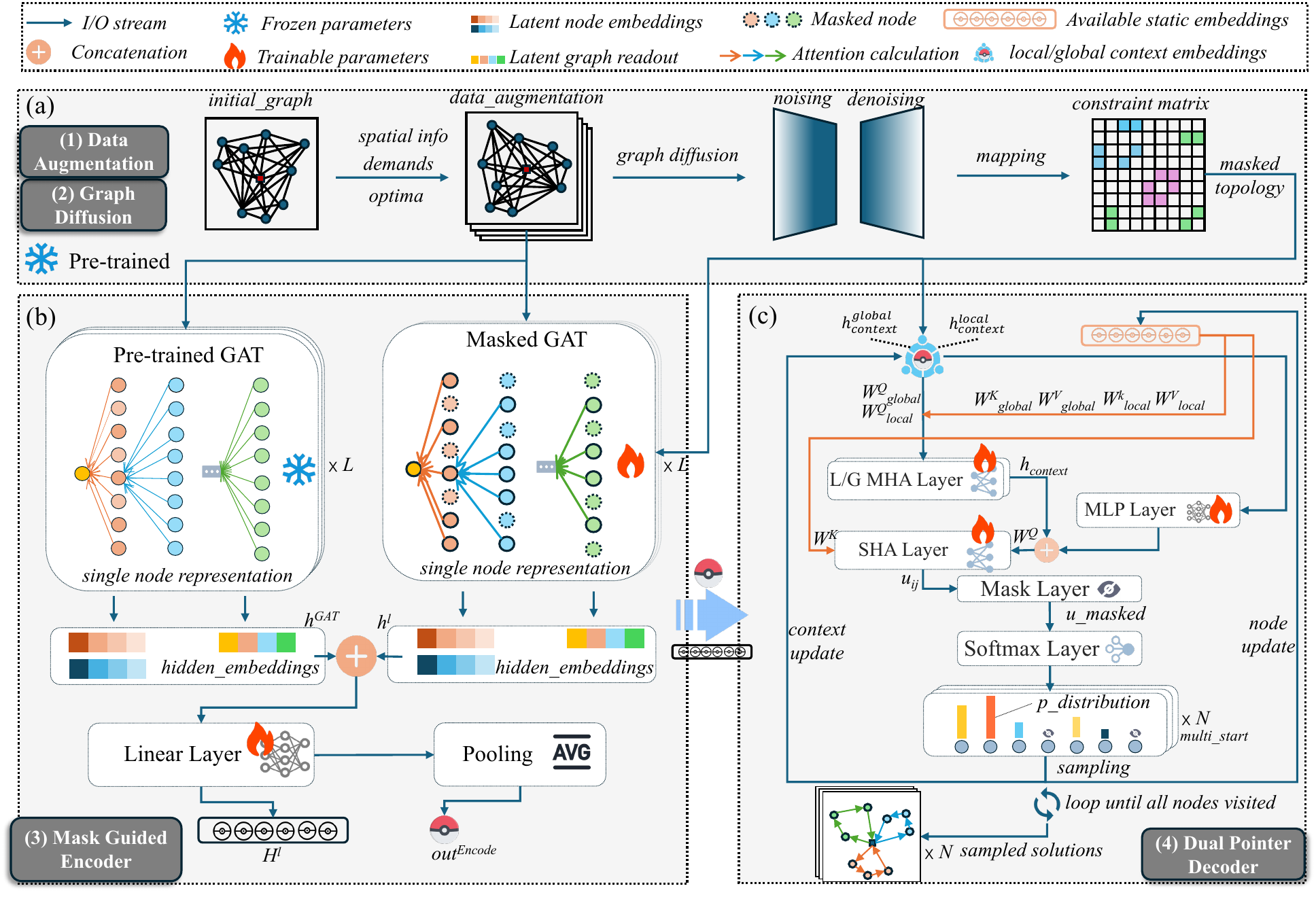}
    \caption{Overall architecture. (a) Graph‑diffusion prior with two submodules: (1) data augmentation and (2) constraint‑matrix generation. (b) Mask‑guided encoder: a pretrained multi‑layer GAT extracts global representations, and a trainable masked GAT extracts local representations; their fusion yields node embeddings. (c) Dual‑pointer decoder: two attention pointers fuse global and local decisions, conditioned on the global graph embedding and the current system state.}
    \label{fig:pipeline}
\end{figure*}
We employ the constraint matrix predicted by the graph diffusion model as a learned topological mask to build a GAT-based encoder for representation learning on unseen data. Compared with fully connected Transformer-style attention, this mask mitigates over-smoothing by restricting message passing to plausible subpath relations, yielding sharper node embeddings and, in turn, more discriminative decoder attention during decoding. The fusion architecture is shown in Fig.~\ref{fig:pipeline}(b)–(c).

\subsubsection{Masked Graph Encoder}
As shown in Fig.~\ref{fig:pipeline}(b), the encoder fuses global and local node representations. The global component is produced by the same pre-trained GNN used in the diffusion model: each instance is treated as a fully connected graph, and attention weights cross-node aggregation to yield the latent embedding $h^{\mathrm{GAT}}$. These parameters remain frozen. The local component is produced by a GAT whose edges are defined by the sparse constraint matrix predicted by the diffusion model. During attention, each node aggregates only over neighbors indicated by its matrix row, masking interactions across different subpaths. Thus, representation learning is restricted to the local sparse neighborhood specified by the constraint matrix.
\begin{equation}\label{attantionscore}
    \alpha_{i j}^{\ell} = \operatorname{Softmax}\left( \frac{\left(\boldsymbol{W}_q^{head} \boldsymbol{h}_i^{(\ell-1)}\right)^\top \boldsymbol{W}_k^{head}\boldsymbol{h}_j^{(\ell-1)} }{\sqrt{d_k}} \right)
\end{equation}
\begin{equation}\label{eq:maskembedding}
    \boldsymbol{h}_i^\ell =  BN \left( \|_{head=1}^{Head}\sum_{j\in\mathbf{\hat{x}}^i_0} \left( \alpha_{ij}^{ (\ell-1)}\boldsymbol{W}_v^{head} \left( \boldsymbol{h}_j^{(\ell-1)}  \right) \right)\oplus \boldsymbol{h}_i^{\ell-1}\right)
\end{equation}
Equation~(\ref{attantionscore}) computes inter-node attention scores, where $\mathbf{W}_q^{head} \in \mathbb{R}^{\frac{d}{k}\times d}$ and $\mathbf{W}_k^{head} \in \mathbb{R}^{\frac{d}{k}\times d}$ represent learnable weight matrices, $head$ denotes the index of multi-head attention heads, and $d_k$ represents the dimension of the latent vector for each attention head. 
Equation~(\ref{eq:maskembedding}) aggregates the multi-head representations of the unmasked node $j$ in the $i-th$ row of the constraint matrix $\mathbf{\hat{x}}_0$ predicted by the graph diffusion network with the current node $i$. $\mathbf{W}_v^{head} \in \mathbb{R}^{\frac{d}{k}\times d}$ represent learnable value-vector weight matrix. The operator $\|$ denotes the concatenation operation, while $\oplus$ represents a residual connection.

The encoder then fuses the graph representation $\textbf{h}^{GAT}\in\mathbb{R}^{N\times d}$ with the masked representation $\textbf{h}^\ell\in\mathbb{R}^{N\times d}$ via a 2-layer MLP, and uses the output of an average-pooling layer as the graph-level encoder output, as in Equation~(\ref{encodeout}).
\begin{equation}\label{encodeout}
    RO^{Encode}= \operatorname{Mean}\left(\operatorname{MLP}(\textbf{h}^\ell+\textbf{h}^{GAT})\right)
\end{equation}
\subsubsection{Dual-Pointer Fusion Decoder}
Similar to prior work~\cite{kwonPOMOPolicyOptimization2020}, our decoder operates autoregressively: given node representations and the encoder’s global graph embedding, a lightweight attention layer computes scores between the current context and candidate nodes to form a distribution for sampling the next visit. During decoding, the vehicle-capacity embedding $MLP(RC_t)$,
where $RC_t$ is the remaining capacity at step $t$, and the representation of the most recently visited node $h_{\tau_{t-1}}^l$ are updated online. Despite its efficiency, this attention can struggle to distinguish similar nodes. We therefore introduce a dual-pointer fusion decoder: a local pointer restricts attention to candidates indicated by the constraint matrix predicted by the graph diffusion model, while a global pointer scores all nodes. Fusing these scores injects a targeted bias, encouraging exploration early and promoting stability later, thereby balancing exploration and exploitation. Fig.~\ref{fig:pipeline}(c) illustrates stepwise parallel decoding and sampling for $N$ start nodes.

The decoder input comprises static and dynamic (context) components. The static input consists of the available node embeddings $\boldsymbol{h}_i^l$ at step $t$, fused with $\boldsymbol{h}_i^{\mathrm{GAT}}$ as in Equation~(\ref{Eq:decode_input}); these remain fixed during decoding because the two encoder GATs already encode decision-relevant information:
\begin{equation}\label{Eq:decode_input}
h_{static}^t =
\mathrm{MLP}(\boldsymbol{h}_i^l + \boldsymbol{h}_i^{\mathrm{GAT}}),\ \forall i \in \mathcal{V}\backslash\{\tau_{1},\dots,\tau_t\}.
\end{equation}
The dynamic context concatenates the remaining capacity, the depot, and the most recently visited node, and serves as the query in attention. Under the dual-pointer design, we form
$h_{context}^{global}=h_0^l+h_{\tau_{t-1}}^l+RO^{Encode}+\mathrm{MLP}(RC_t)$ and
$h_{context}^{local}=h_0^l+h_{\tau_{t-1}}^l+\mathrm{MLP}(RC_t)$. At each step, keys are drawn only from unvisited nodes, shrinking the search space without extra cost.

To exploit the diffusion prior, we apply the constraint matrix as an attention mask (as in Equation~(\ref{eq:maskembedding})): the local pointer attends within the masked neighborhood, while the global pointer attends over all nodes. Their multi-head attention outputs are 
\begin{equation}\label{eq:decodemha}
    L/G MHA = \left\{
    \begin{array}{ll}
       MHA(h_{context}^{b},h_{static}^t,h_{static}^t) & b=global \\
        MHA(h_{context}^{b}, h_{static}^t,h_{static}^t) & b=local
    \end{array}
    \right.
\end{equation}
and are fused to form a novel context vector (Equation~(\ref{hcontext})):
\begin{equation}\label{hcontext}
\begin{aligned}
    h_{context} &= \sum_{b=local}^{global}MHA(h_{context}^{b},h_{static}^t,h_{static}^t)\\ 
    &=||_{head}\sum_{j\in\hat{x}_0^i}u_{i,j}^{local}W^V_{local}h_j\\ 
    &+ ||_{head} \sum_{j\notin\{\tau_{0}...\tau_{t-1}\}}u_{i,j}^{global}W_{global}^Vh_j
\end{aligned}
\end{equation}
This vector integrates rich global and local information, thereby enabling effective single-step decision-making. The dual pointers independently compute their respective attention scores $u_{i,j}^{local}$,$u_{i,j}^{global}$ and update the value matrices associated with their corresponding available node embeddings $h_j$, as detailed in Equations~(\ref{eq:uijlocal}) and (\ref{eq:uijglobal}). 
\begin{equation}\label{eq:uijlocal}
    u_{i,j} ^{local} = Softmax\left(\frac{(W^Q_{local}h_{context}^{local})^TW^K_{local}h_{static}^t}{\sqrt{d_{head}}}\right) 
\end{equation}

\begin{equation}\label{eq:uijglobal}
    u_{i,j} ^{global} = Softmax\left(\frac{(W^Q_{global}h_{context}^{global})^TW^K_{global}h_{static}^t}{\sqrt{d_{head}}}\right) 
\end{equation}
In essence, the multi-head attention layer delineated in Equations~(\ref{hcontext})-(\ref{eq:uijglobal}) supplies two distinct sets of query-key-value (QKV) weights, denoted as \( W_{local}^Q \),\( W_{gloval}^Q \), \( W_{local}^K \), \( W_{global}^K \),\( W_{local}^V \), and \( W_{global}^V \) all $\in\mathbb{R}^{\frac{d}{k}\times d}$, which enable the learning and updating processes to generate an intermediate context vector \( h_{contxt} \) that effectively combines global and local information. Here, the operator \( \| \) denotes the concatenation of the multi-head value matrices. For brevity, superscripts $k \in[1,head]$ designating the attention heads have been omitted in the equations. This fusion yields two benefits: nodes highlighted by both pointers act as anchors that stabilize exploration, and when global attention is diffuse (e.g., clustered features), the model prioritizes constraint-consistent local nodes, improving robustness to look-alike candidates.

In previous studies\cite{koolAttentionLearnSolve2019,kwonPOMOPolicyOptimization2020,wang2024gase,leiSolveRoutingProblems2022,luo2023neural}, attention weights are computed from the context vector and available nodes and then used to update only the nodes’ value matrices, leaving the context vector unchanged during learning. To better exploit contextual information, we introduce an auxiliary perception layer that processes the current-step local context and fuses it with vectors from the local and global attention mechanisms, as detailed in Equation~(\ref{eq:hbar}).
\begin{equation}\label{eq:hbar}
    \bar{h}_{context} = MLP(h_{context}^{local})+h_{context}
\end{equation}
The resultant vector \(\bar{h}_{context}\) serves as the updated context for computing attention scores via a single-head attention layer. The probability distribution \(p_{\theta}\) is then derived through hyperbolic tangent activation ($tanh$) and clipping constant $C$, as formulated in Equations~(\ref{eq:shau}) and (\ref{eq:ptheta}).
\begin{equation}\label{eq:shau}
\begin{aligned}
    u_{i,j} &= SHA(\bar{h}_{context},h_{static}^t)\\ &=\left(\frac{(W^Q\bar{h}_{context})^TW^Kh_{static}^t}{\sqrt{d}}+log(Saving_{i,j})\right)
\end{aligned} 
\end{equation}

\begin{equation}\label{eq:ptheta}
    p_{\theta}(\tau_t|\tau_0...\tau_{t-1}) = Softmax\left(C\cdot tanh(u_{\tau_{t-1},\tau_t}))\right)
\end{equation}
Notably, when computing attention weights in Equation~(\ref{eq:shau}), we incorporate a heuristic savings term \(Saving_{i,j} =\text{dist}(0,i) + \text{dist}(0,j) - \text{dist}(i,j)\)(with 0 denoting the depot), which measures the distance saved by visiting $i$ and $j$ consecutively rather than separately. Drawing upon the established mileage savings algorithm\cite{segerstedt2014simple}, this term biases exploration toward nodes that offer greater economic efficiency.

\subsection{Training and Inference}
\subsubsection{Training}
As shown in Fig.~\ref{fig:pipeline}, we first pre-train the graph diffusion model via standard supervised learning on synthetic instances from POMO~\cite{kwonPOMOPolicyOptimization2020}. Labels are generated by HGS on 50,000 instances, yielding the corresponding adjacency and constraint matrices. We then freeze the diffusion model and train the encoder–decoder with unsupervised reinforcement learning, using it to iteratively construct solutions from a sketch. This section details the RL procedure.

During training, we use Monte Carlo estimates over batched trajectories to optimize the expected reward, \( \mathcal{L} ({\pi_\theta}|\mathcal{G}) = \mathbb{E}_{\tau\sim p_{\pi_\theta}(\tau | \mathcal{G})}\mathcal{R}(\tau)\). To promote more stable model convergence, we adopt the REINFORCE \cite{zhang2021sample}, with a self-critic baseline, maximizing reward (equivalently, minimizing tour length).  We further introduce an enhancement baseline, evaluating two strategies: (i) the average tour length from multi-start sampling over several starting nodes, and (ii) the average tour length from single-start greedy inference. Empirically, the single-start baseline converges faster but is prone to local optima; hence we use the multi-start baseline. The comprehensive training algorithm is outlined as follows:
\begin{algorithm}
	\renewcommand{\algorithmicrequire}{\textbf{Input:}}
	\renewcommand{\algorithmicensure}{\textbf{Output:}}
    \caption{Self-critic Reinforce Training}\label{alg1}
\begin{algorithmic}[1]
    \REQUIRE Training set $D_T$; policy network $\pi_\theta$; Training epochs $E$; Batch size $B$; Steps per epoch $T$; Number of starting nodes $N$; Optimizer $Adam$.
    \STATE Initialization: $\theta$
    \FOR{ $e$ in $1...E$}
    \FOR{ $t$ in $1...T$}
    
    \STATE $\mathcal{G}_i \gets RandomInstance(D_T), \forall i \in \{1,....,B\}$
    \STATE $\mathcal{S}_{\{s_i^1,...,s_i^N\}} \gets NStart(\mathcal{G}_i),\forall i \in \{1,....,B\}$
    \STATE $\tau_i^n \gets SampleSolution(\pi_\theta,\mathcal{S})$
    \STATE Compute $\mathcal{R}(\tau_i^n) $ through Eq. (\ref{rewardcal})
    \STATE $\mathcal{R}(\tau_i^b)=\frac{1}{N}\sum_{n=1}^N \mathcal{R}(\tau_i^n) $
    \STATE $\nabla_\theta\mathcal{L}(\theta) \gets\frac{1}{B\times N} \sum_{i=1}^B\sum_{n=1}^N (\mathcal{R}(\tau_i^n) - \mathcal{R}(\tau_i^b))\nabla_\theta log p_{\theta}(\tau_i^n | \mathcal{S})$
    \STATE $\theta \gets Adam(\theta, \nabla_\theta\mathcal{L}({\theta})$
    \ENDFOR    
    \ENDFOR
    \ENSURE Parameters set $\theta$ of actor network       
\end{algorithmic}  
\end{algorithm}
As in POMO~\cite{kwonPOMOPolicyOptimization2020}, the \(\textit{N}\text{\textit{start}}\) strategy uses all nodes as starting points for instances with \(N\leq 100\) to enable parallel decoding; for \(N>100\), it selects the 100 nodes closest to the depot. This caps start-node parallelism at a tractable scale and aligns with the observation that, in large instances, both the first and last nodes (the latter in the reverse tour) are typically near the depot. During training, actions are sampled from the policy \(\pi_\theta\) via its distribution \(p_\theta\), enlarging the exploration space in action selection. The final output is the converged policy parameters, obtained once rewards stabilize.

\subsubsection{Inference}
At inference, unlike training, we exploit the permutation invariance of optimal solutions in geometric space to apply data augmentation under the current policy network parameters and test whether augmented instances yield better tours. As delineated in Algorithm~\ref{alg:inference}, a single instance is augmented by a dedicated routine that applies geometric symmetries to generate total 
$A$ variants. Notably, this uses the same geometric space as the diffusion-stage augmentation but omits demand-level augmentation. Because the optimal solution is unknown at inference, demand-constraint validity cannot be verified, so demands are not modified.
\begin{algorithm}
	\renewcommand{\algorithmicrequire}{\textbf{Input:}}
	\renewcommand{\algorithmicensure}{\textbf{Output:}}
    \caption{Inference}\label{alg:inference}
\begin{algorithmic}[1]
    \REQUIRE Instance $\mathcal{G}$; policy network $\pi_\theta$; Number of starting nodes $N$; Number of augmented instances $A$.

    \STATE $\{\mathcal{G}_1,...,\mathcal{G}_A\} \gets AugmentedInstance(\mathcal{G})$
    \STATE $\mathcal{S}_{\{s_i^1,...,s_i^N\}} \gets NStart(\mathcal{G}_i),\forall i \in \{1,....,A\}$
    \STATE $\tau_i^a \gets GreedyRulloutSolution(\pi_\theta,\mathcal{S}), \forall a\in \{1,...,A\},\forall i \in\{1,...,N\}$
    \STATE $a_{max},i_{max} \gets argmax_{a,i}\mathcal{R}(\tau_i^a)$
    \ENSURE $\tau_{i_{max}}^{a_{max}}$       
\end{algorithmic}  
\end{algorithm}

\section{Experiments}
We evaluate the proposed graph diffusion–augmented generative solver through three key research questions: RQ1) performance on public benchmarks; RQ2) zero-shot generalization to out-of-distribution (OOD) instances; and RQ3) sensitivity to hyperparameters and ablation settings, including the removal of the diffusion prior from the graph-mask–guided encoder and the dual-pointer modules.
\subsection{Experiment Setup}
\subsubsection{Datasets}
The datasets consist of synthetic instances and public benchmarks. 

To train the graph diffusion model, we generate 50{,}000 labeled instances with solutions computed by HGS to balance efficiency and accuracy. For RL of the autoregressive policy, we synthesize 200{,}000 unlabeled instances from the same distribution. Node coordinates are drawn uniformly from [0,1], demands uniformly from 1 to 9, depot locations are random, and vehicle capacities are set to 30/40/50 for problem sizes of 20/50/100, respectively.

For public cross-distribution benchmark evaluation, we employ the CVRPLIB XML100 suite~\cite{queiroga2022}, comprising 10,000 test instances. The suite categorizes instances along four attributes: node distribution, depot distribution, demand distribution, and route length, resulting in 378 combinatorial groups. To the best of our knowledge, this work presents the first systematic evaluation of constructive heuristic neural solvers on XML100. 
\subsubsection{Parameters and Environmental Setup} Table~\ref{tab:para} outlines the key parameters for the graph diffusion and autoregressive models. The diffusion model uses 1,000 denoising steps during training; however, following Denoising Diffusion Implicit Models (DDIM), far fewer steps are used at inference. We conduct a grid search over inference step counts and per-step sample sizes, evaluating AUC and runtime; results appear in the parameter‑sensitivity section. All experiments were run on an NVIDIA RTX A6000 (48 GB) and an Intel Xeon Silver 4126 CPU.
\begin{table}[htb]
\caption{Experimental parameters}\label{tab:para}
    \centering
    \resizebox{\linewidth}{!}{
    \begin{tabular}{l|c}
    \Xhline{1pt}
    Parameters & Values\\
    \hline
       Diffusion Training steps $T$  & 1000 \\
       Diffusion Inference steps $T^\prime$& 50 \\
       Diffusion linear noise schedule $\beta_1$ & $10^{-4}$\\
       Diffusion linear noise schedule $\beta_T$ & $0.02$\\
       Diffusion batch size & 32\\
       Diffusion Training epoch & 50 \\
      \hline
       Pre-trained GAT layers $L^\prime$ & 5\\
       Pre-trained GAT batch size & 64 \\
       Pre-trained GAT embedding dim & 128 \\
       Pre-trained GAT attention head number & 8 \\
      \hline
       Encoder graph layers $L$& 5\\
       Decoder attention head number $head$ & 8\\
       Decoder QKV dim &16\\
       Logit clipping $C$ & 10\\
       Learning rate $\alpha$ &$10^{-4}$\\
       Weight decay & $10^{-6}$ \\
       Optimizer & Adam \\
       Reinforce batch size $B$ & 32 \\
       Hidden dim $d$ & 128\\
       Training Epoch $E$ & 100 with early stop \\
       \Xhline{1pt}
        
    \end{tabular}
  }
\end{table}
\subsubsection{Baselines}
We survey representative methods and group them into four categories: general solvers (Gurobi, LKH3, HGS, OR-Tools); sequence-based autoregressive solvers (Ptr-Net); fully connected Transformer-based autoregressive solvers (AM, POMO, LEHD, ReLD); and graph-structured autoregressive solvers (E-GAT, GASE).  Notably, no diffusion model-based solvers directly address CVRP, largely due to its complex constraints and heavy reliance on posterior search (e.g., MCTS). The predominant approach entails generating a probability heatmap followed by search-based decoding.  Most diffusion approaches generate a probability heatmap followed by search-based decoding. Although neighborhood-search refinement of an initial feasible solution can yield strong results, it is computationally expensive and ill-suited for online use, and it underutilizes the intrinsic value of the neural policy’s solutions. We therefore exclude such methods from comparison. The selected neural network solver baselines are delineated as follows:
\begin{itemize}
\item  Pointer Networks (PtrNets)~\cite{nazariReinforcementLearningSolving2018}: Pioneers reinforcement learning with LSTMs for CVRP and TSP, achieving strong performance.

\item Attention Model (AM)~\cite{koolAttentionLearnSolve2019}: The first to couple Transformers with reinforcement learning, fully exploiting attention in both encoder and decoder; it serves as the backbone for many subsequent autoregressive methods.

\item  POMO~\cite{kwonPOMOPolicyOptimization2020}: An AM-based autoregressive RL model introducing multi-start parallel sampling and inference-time data augmentation.

\item  LEHD~\cite{luo2023neural}: A supervised approach with a lightweight encoder and a heavy decoder (stacked multi-attention layers), plus a sampling-and-reconstruction strategy for accuracy and scalability.

\item ReLD~\cite{huang2025rethinking}: Emphasizes encoding over decoding; augments AM/POMO backbones with a feedforward layer to capture the decoding state and better exploit the current system state.

\item ELG~\cite{gao2023towardslocalglobal}: Builds on POMO with combined global–local attention; uses KNN in polar coordinates to add an out-of-local penalty in the RL loss, focusing decisions on nearby nodes while preserving global awareness.

\item  E-GAT~\cite{leiSolveRoutingProblems2022}: The first RL autoregressive model to fully leverage GAT for encoding, incorporating node distances as edge features to improve representations for downstream decoding.

\item  GASE~\cite{wang2024gase}: Extends E-GAT with a sampling strategy in the GAT encoder to enhance representational accuracy, achieving state-of-the-art performance among graph-structured autoregressive models.

\end{itemize}
We compare against these neural baselines on the test set to address the three research questions and provide detailed analysis of our model.
\subsection{Experimental Results}
\subsubsection{Performance Analysis on Synthetic and CVRPLIB Benchmark}
\begin{table*}[htb]
\caption{Comparison with SOTA methods on randomly generated CVRP instances. “–” in the Gurobi row indicates no result within the time limit; Gurobi, HGS, and LKH3 results are reproduced from prior studies~\cite{koolAttentionLearnSolve2019,kwonPOMOPolicyOptimization2020,leiSolveRoutingProblems2022}. “*” in the Time column denotes the total runtime of a single-threaded, sequential OR-Tools run with a 1s per‑instance search limit over 1,000 test cases. Entries in bold and underlined show the best results across neural solvers }
        \label{tab:basicresult}
	\centering
    \resizebox{\linewidth}{!}{
	\begin{tabular}{l|ccc|ccc|ccc}
		\Xhline{1pt}
		\multirow{2}*{Model} & \multicolumn{3}{c|}{CVRP20} &  \multicolumn{3}{c|}{CVRP50}& \multicolumn{3}{c}{CVRP100}\\ 
		& Objective & Gap(\%) & Time & Objective & Gap(\%) & Time & Objective & Gap(\%) & Time\\
	
        \hline 
		Gurobi\cite{Gurobi2024}       & 6.10 & 0.00 & 13h  & -     & -    & -  &   -   &  -   &  -  \\
        HGS\cite{kwonPOMOPolicyOptimization2020}         & 6.11 & 0.16 &  8.2h &  10.36&  0.00 &    21h &  15.51 & 0.00  &32h  \\
		LKH3\cite{helsgaun2017extension}          & 6.14 & 0.65 & 2h & 10.38 & 0.19 & 7h & 15.65 & 0.90 & 13h  \\
		OR Tools\cite{ortools}     & 6.18 & 1.31 & *1000s  & 10.98 & 5.98 & *1016s  & 17.09 & 10.19 &  *1054s  \\
        \hline
            PtrNet [2018]       & 6.59 & 8.03 & $<$1s  & 11.39 & 9.94 & $<$1s  & 17.23 & 11.09 & $<$1s \\
           
		AM [2019]     & 6.40 & 4.97 & 1s  & 10.98 & 5.98 & 3s & 16.80 & 8.32  & 8s  \\
        AM Sampling [2019]     & 6.24 & 2.30 & 3m  & 10.59 & 2.22 & 7m & 16.16 & 4.19  & 30m  \\
            
            POMO [2020]         &6.35& 4.09 &  1s&  10.74&  3.67 &    1s &  16.15 & 4.13  &3s  \\
            POMO$\times8Aug$ [2020]         & \textbf{\underline{6.14}} & \textbf{\underline{0.65}} &  5s&  10.42&  0.58 &    26s &  15.73 & 1.42  &2m  \\
            LEHD Greedy [2023]         & 6.36 & 4.26 &  2s&  10.92&  5.41 &    2s &  16.22 & 4.59  &30s  \\
             ELG$\times8Aug$ ensemble 0.4 Local Size [2024]         & 6.35 & 4.16 &  4s&  10.47&  1.06 &    10s &  15.84 & 2.13  &3.18m  \\
            ReLD $\times 8Aug$[2025]         & 6.15 & 0.89 &  3s&  10.43& 0.66  &   9s &  15.80 & 1.87  &1.34m  \\
            \hline
           
		E-GAT [2023]        & 6.26 & 2.60 & 2s & 10.80  & 4.25 & 7s & 16.69 & 6.68 & 17s \\
        GASE [2024]       & 6.21 & 1.80 & 2s &10.74  & 3.52 & 3s & 16.37  & 4.60   &  7s \\
        \hline 
		{\bf Our approach} &  6.20 & 1.64  & 2s  &  10.64 & 2.70  &  5s  & 15.76 & 1.61 & 12s \\
         {\bf Our approach $\times 8Aug$}         & \textbf{\underline{6.14}} & \textbf{\underline{0.65}} &  6s&  \textbf{\underline{10.41}}&  \textbf{\underline{0.48}} &    30s & \textbf{\underline{15.70}}  & \textbf{\underline{1.23}}  &2m \\
		\Xhline{1pt}
	\end{tabular}
}
\end{table*}
We first evaluate our model on 1{,}000 synthetic test instances whose node coordinates and demands follow the training distribution, with random depot locations. Table~\ref{tab:basicresult} reports results: the first block lists general-purpose solvers using exact (for small scales) or heuristic methods; the second block lists autoregressive neural solvers, split into graph‑structured and non‑graph‑structured variants. The \textit{Objective} column reports the average solution distance, while the \textit{Gap} column quantifies the relative deviation from the current best baseline, computed as \((Obj - Obj_{best}) / Obj_{best}\). 

Evidently, learning-based approaches exhibit substantial advantages in solution time and generalization. Once trained,  they produce near‑optimal in‑distribution solutions with low latency. Because LEHD, ELG, and ReLD target large-scale settings and release models only for 100-node cases, we adapted the authors’ code under matched conditions to evaluate smaller scales.  On the synthetic benchmark in Table~\ref{tab:basicresult}, our method attains SOTA performance among recent representative learning-based solvers, particularly without augmented inference, and demonstrates a modest efficiency edge. We attribute these gains to the graph diffusion prior, which captures constraint patterns via geometric and demand augmentation, thereby strengthening the mask‑guided encoder and the dual‑pointer decoder that fuses local and global attention.

Furthermore, to guard against overfitting to the synthetic distribution, we further evaluate on CVRPLIB across multiple scales, comparing against leading models trained on synthetic data. The results are presented in Table~\ref{tab:vrplibresult}. Under distribution shift, autoregressive models exhibit an average gap of over 5\% to optimal, markedly worse than in‑distribution. Moreover, our experiments underscore the limited scalability of autoregressive models with lightweight decoders (i.e., those incorporating few attention layers in the decoder). In particular, deploying a model pre-trained on 100-node instances to problems with 200 or more nodes yields pronounced performance declines. A key factor is per‑step out‑of‑distribution inputs in large instances, which raise stepwise error rates; as horizons lengthen, errors accumulate. Besides, policy‑gradient RL with trajectory‑level rewards scales poorly as the per‑step action space grows combinatorially, yielding sparse signals and slow convergence. Despite these challenges, our approach outperforms contemporaries using comparable training regimes (i.e. light decoder models such as POMO). Additionally, it exhibits a lower standard deviation than the heavy decoder models (LEHD).  

\begin{table*}[!htb]
\caption{Model performance on CVRPLIB across multiple problem scales. Boldface upon underlines denotes the best result among compared methods; “–” indicates omitted (not reported) entries. The Gap column quantifies the discrepancy from the optimal solution.}\label{tab:vrplibresult}
	\centering
        \resizebox{\linewidth}{!}{
	\begin{tabular}{l|c|c|cc|cc|cc|cc}
		\Xhline{1pt}
		\multirow{2}*{Data instance} & \multirow{2}*{Nodes number}& \multicolumn{1}{c|}{Optimal Solution} &  \multicolumn{2}{c|}{Our Approach$\times8$Aug}& \multicolumn{2}{c|}{E-GAT}&\multicolumn{2}{c|}{POMO$\times8$Aug}&\multicolumn{2}{c}{LEHD}\\ 
		
		& & Objective & Objective & Gap(\%) & Objective & Gap(\%) &  Objective & Gap(\%) &Objective & Gap(\%)   \\
		\hline 
 
            A-n44-k6  & 43  & 937   & \textbf{\underline{972}}  & \textbf{\underline{3.74}}   & 979     & 4.49  & 997   & 6.40 & 1000 & 6.74 \\
		A-n45-k6  & 44  & 944   & \textbf{970}  & \textbf{2.75}   & 984     & 4.23  & 993  & 5.22 & 994 & 5.33 \\
           
		A-n63-k10 & 62  & 1314  & 1337  & 1.75    & 1519    & 15.60 & \textbf{\underline{1329}}  & \textbf{\underline{1.16}} & 1463 & 11.32 \\
            A-n64-k9  & 63  & 1401  & 1493 & 6.57   & 1659    & 18.40 & \textbf{\underline{1430}}  & \textbf{\underline{2.10}} & 1478 & 5.55 \\
            A-n69-k9  & 68  & 1159  & \textbf{\underline{1193}}  &  \textbf{\underline{2.93}}   & 1264    & 9.05  & 1199  & 3.47 & 1264 & 9.04\\
            B-n34-k5  & 33  & 788   & \textbf{\underline{811}}   & \textbf{\underline{2.91}}    & 812     & 3.04  & 903   & 14.66 &813&3.23 \\
            B-n35-k5  & 34  & 955   & \textbf{\underline{965}}   &  \textbf{\underline{1.05}}   & 986     & 3.24  &1005  & 5.28 & 976 & 2.20 \\
            B-n45-k6  & 43  & 678   & \textbf{\underline{710}}   &  \textbf{\underline{4.72}}  & 729     & 7.52  & 737   & 8.73 & 739 & 8.97\\
            B-n51-k7  & 50  & 1032  & 1039  & 0.68   & 1045    & 1.25  & 1088  & 5.48 & \textbf{\underline{1034}} & \textbf{\underline{0.25}}\\
            P-n50-k8  & 49  & 631   & 655   & 3.80    & 655     & 3.80  & 661   & 4.83 & \textbf{\underline{654}} & \textbf{\underline{3.65}} \\
            P-n51-k10 & 50  & 741   & 759   & 2.43   & 811     & 9.44  & 759  & 2.43 & \textbf{\underline{754}} & \textbf{\underline{1.70}} \\
            P-n70-k10 & 69  & 827   & 872   &   5.44  & 865     & 4.59 & 867   & 4.79 & \textbf{\underline{857}} & \textbf{\underline{3.60}} \\
            E-n101-k14  & 100  & 1067   &  1133  &  6.18  & 1163     & 8.96  & \textbf{\underline{1116}}   & \textbf{\underline{4.22}} & 1157 & 8.41 \\
  M-n101-k10  & 100  & 820   &  831  & 1.34   &  873     & 6.46  & 842   & 2.68 & \textbf{\underline{828}} & \textbf{\underline{1.03}} \\
  M-n121-k7  & 120  & 1034  & 1141 & 10.35  & 1144    & 10.63 & \textbf{\underline{1110}}   & \textbf{\underline{7.35}} & 1206 & 16.60\\

  M-n151-k12 &	150	& 1015	&1153 & 13.60 &	1106&	8.96&	1076&	6.01& \textbf{\underline{1070}} & \textbf{\underline{5.44}} \\
  M-200-k16 & 199 &1274 & 1484 & 16.48 & 1438 & 12.87 & 1374 & 7.85  & \textbf{\underline{1354}} & \textbf{\underline{6.24}}\\  
            \hline
		{Average Gap}  & - & - & - &\textbf{\underline{5.10}} & - & 6.49  & - & 5.45 &-& 5.84 \\
		\Xhline{1pt}
	\end{tabular}
 }
\end{table*}
To summarize, compared with state‑of‑the‑art models, our approach yields superior solution quality and competitive inference speed on in‑distribution data, and generalizes better across scales for problems with 100 nodes or fewer.
\subsubsection{Few-Shot OOD Analysis}
To further substantiate and investigate the model's generalization capabilities on out-of-distribution (OOD) feature data, we conducted experiments and detailed comparative analyses on the CVRP XML100 dataset using identical architectures. The XML100 dataset encompasses 10,000 instances, each with a problem scale of 100 nodes, yet featuring intricate and diverse feature distributions. The four categorization dimensions: depot location, node coordinate distribution, demand slackness distribution, and route length distribution, yield a total of 19 classes and 378 combinations. Each instance is characterized by a complex combination across the four distributional dimensions. The experimental results are presented in Table~\ref{tab:xml100result}.
\begin{table*}[!htb]
\caption{Comparative analysis of out‑of‑distribution (OOD) generalization. Boldface upon underlines indicates that the current subcategory achieves a lower average gap or lower average standard deviation than the competing model without/with $8$ times data augmentation. Inference uses a pretrained 100‑node model trained on synthetic data.}\label{tab:xml100result}
	\centering
        \resizebox{\linewidth}{!}{
	\begin{tabular}{c|c|c|cc|cc||cc|cc}
		\Xhline{1pt}
		\multirow{2}{*}{Dimension} & \multirow{2}{*}{Description}& \multirow{2}{*}{Number of Instance} & \multicolumn{2}{c|}{POMO} &\multicolumn{2}{c||}{Our Approach} & \multicolumn{2}{c|}{POMO$\times8$Aug} & \multicolumn{2}{c}{Our Approach$\times$ 8 Aug} \\ 
        \multirow{2}{*}{}&\multirow{2}{*}{}&\multirow{2}{*}{}&Gap(\%)&Gap Std&Gap(\%)&Gap Std&Gap(\%)&Gap Std&Gap(\%)&Gap Std\\
		\hline
        \multirow{3}{*}{Depot Distribution}&random & 3402 & 8.80 & 7.10 & \textbf{\underline{8.27}} & \textbf{\underline{6.42}} & 5.86 & 4.25 & \textbf{\underline{5.62}} & \textbf{\underline{3.87}}\\
        \multirow{3}{*}{}&centered & 3322 & 10.60 & 9.67 & \textbf{\underline{9.95}} & \textbf{\underline{8.13}}& 6.59 & 4.62 & \textbf{\underline{6.27}} & \textbf{\underline{4.08}}  \\
        \multirow{3}{*}{}&cornered & 3276 & 8.11 & 6.60 & \textbf{\underline{7.97}} & \textbf{\underline{6.37}} & 5.50 & 4.17 & \textbf{\underline{5.40}} & \textbf{\underline{4.05}}  \\
		\hline 
        \multirow{3}{*}{Customer Distribution}&random & 3360 & 8.28 & 6.59 & \textbf{\underline{7.86}} & \textbf{\underline{6.25}} & 5.58 & 3.84 & \textbf{\underline{5.36}} & \textbf{\underline{3.56}}  \\
        \multirow{3}{*}{}&clustered & 3322 & 10.59 & 9.11 & \textbf{\underline{9.98}} & \textbf{\underline{8.07}} & 6.62 & 5.05 & \textbf{\underline{6.42}} & \textbf{\underline{4.66}}  \\
        \multirow{3}{*}{}&random-clustered & 3318 & 8.75 & 7.83 & \textbf{\underline{8.36}} & \textbf{\underline{6.59}} & 5.77 & 4.06 & \textbf{\underline{5.52}} & \textbf{\underline{3.66}} \\            
		\hline
        \multirow{7}{*}{Demands Distribution}&unitary & 1432 & 14.11 & 9.67 & \textbf{\underline{11.89}} & \textbf{\underline{8.46}} & 8.41 & 5.50 & \textbf{\underline{7.25}} & \textbf{\underline{4.71}} \\
        \multirow{7}{*}{}&small values large CV & 1428 & \textbf{\underline{6.59}} & 6.19 & 6.61 & \textbf{\underline{6.02}} & \textbf{\underline{4.46}} & \textbf{\underline{3.63}} & 4.55 & \textbf{\underline{3.63}} \\
        \multirow{7}{*}{}&small values small CV & 1428 & 8.14 & 8.43 & \textbf{\underline{7.74}} & \textbf{\underline{6.41}} & 5.54 & 4.63 & \textbf{\underline{5.30}} & \textbf{\underline{4.25}} \\
        \multirow{7}{*}{}&large values large CV & 1428 & \textbf{\underline{6.73}} & \textbf{\underline{5.44}} & 7.07 & 5.73 & \textbf{\underline{4.79}} & \textbf{\underline{3.54}} & 4.80 & 3.55 \\
        \multirow{7}{*}{}&large values small CV & 1428 & 8.46 & 7.30 & \textbf{\underline{8.07}} & \textbf{\underline{6.73}} & 5.82 & 4.61 & \textbf{\underline{5.50}} & \textbf{\underline{4.06}}\\
        \multirow{7}{*}{}&depending on quadrant & 1428 & \textbf{\underline{6.91}} & 5.41 & 7.01 & \textbf{\underline{5.19}} & \textbf{\underline{4.75}} & \textbf{\underline{3.19}} & 4.90 & 3.30 \\
        \multirow{7}{*}{}&many small \& few large & 1428 & 13.24 & 8.34 & \textbf{\underline{12.77}} & 7.62 & 8.13 & 3.00 & \textbf{\underline{8.06}} & \textbf{\underline{2.96}} \\
        \hline
        \multirow{6}{*}{Vehicle Route Distribution}&very short & 1667 & 9.18 & 7.12 &\textbf{\underline{8.68}} & \textbf{\underline{7.06}} & 5.91 & \textbf{\underline{2.34}} & \textbf{\underline{5.63}} & \textbf{\underline{2.34}} \\
        \multirow{6}{*}{}& short & 1667 & 6.95 & 5.83 & \textbf{\underline{6.54}} & \textbf{\underline{5.15}} & 4.53 & 2.92 & \textbf{\underline{4.36}} & \textbf{\underline{2.72}} \\
        \multirow{6}{*}{}& medium & 1667 & 5.21 & 5.54 & \textbf{\underline{5.09}} & \textbf{\underline{5.15}} & \textbf{\underline{3.00}} & 2.34 & 3.03 & \textbf{\underline{2.26}} \\
        \multirow{6}{*}{}& long & 1667 & 6.14 & 5.49 & \textbf{\underline{5.86}} & \textbf{\underline{4.57}} & 3.76 & 2.64 & \textbf{\underline{3.64}} & \textbf{\underline{2.18}} \\
        \multirow{6}{*}{}& very long & 1666 & 10.04 & 6.73 & \textbf{\underline{9.53}} & \textbf{\underline{5.53}} & 6.54 & 3.40 & \textbf{\underline{6.38}} & \textbf{\underline{2.94}} \\
        \multirow{6}{*}{}& ultra long & 1666 & 17.50 & 9.33 & \textbf{\underline{16.69}} & \textbf{\underline{7.45}} & 12.18 & 4.68 & \textbf{\underline{11.56}} & \textbf{\underline{4.18}} \\
        \Xhline{0.9pt}
        \multirow{2}{*}{} & \multirow{2}{*}{} & \multirow{2}{*}{\textbf{Average}} & \multirow{2}{*}{9.17} & \multirow{2}{*}{7.98} & \multirow{2}{*}{\textbf{\underline{8.73}}} & \multirow{2}{*}{\textbf{\underline{7.07}}} & \multirow{2}{*}{5.99} & \multirow{2}{*}{4.37} & \multirow{2}{*}{\textbf{\underline{5.77}}} & \multirow{2}{*}{\textbf{\underline{4.02}}} \\
        \multirow{2}{*}{} & \multirow{2}{*}{} & \multirow{2}{*}{} &\multirow{2}{*}{}  & \multirow{2}{*}{} &\multirow{2}{*}{}  & \multirow{2}{*}{} & \multirow{2}{*}{} & \multirow{2}{*}{} & \multirow{2}{*}{} & \multirow{2}{*}{} \\

         \Xhline{1pt}
        
	\end{tabular}
 }
\end{table*}
The results demonstrate that, when confronted with instances of equivalent scale but intricate feature distributions, our model exhibits an average gap of approximately 5.77\% relative to the optimal solution. Compared to the current SOTA, it achieves smaller gaps (Gap\%) and lower variability (Gap Std), highlighting improved robustness from the graph‑diffusion prior. However, performance degrades markedly in scenarios involving extreme (long or short) route lengths, uniform demands, clustered customer nodes, and centrally located depots. We attribute this to the encoder's propensity for over-representation and oversmoothing of node embeddings, which rely on coordinates and demands as inputs, yielding diffuse decoder attention and weaker greedy rollouts on OOD data. Our diffusion‑mask mitigates this by constraining aggregation, improving averages by 1.2\% in clustered, uniform‑demand, and extreme‑route cases.

Furthermore, we observed that our model underperforms POMO in scenarios where node demand and vehicle capacity constraints are relatively relaxed, specifically, when node demands are comparatively small but with greater vehicle capacity. This suggests that our model is better suited for instances involving tighter capacity constraints. We attribute this phenomenon to the latent constraint pattern learned by the diffusion prior, which decreases the susceptibility of node representations to oversmoothing when facing tighter constraints. Moreover, longer vehicle routes result in denser constraint matrices, which pose greater challenges for graph diffusion networks in accurate generation compared to sparse counterparts. Additionally, an excessively dense OOD constraint matrix, although upon masking, can still substantially dissolve the decision network's attention computations, thereby amplifying errors.

\begin{figure}[!htb]
    \centering
    \includegraphics[width=1.0\linewidth]{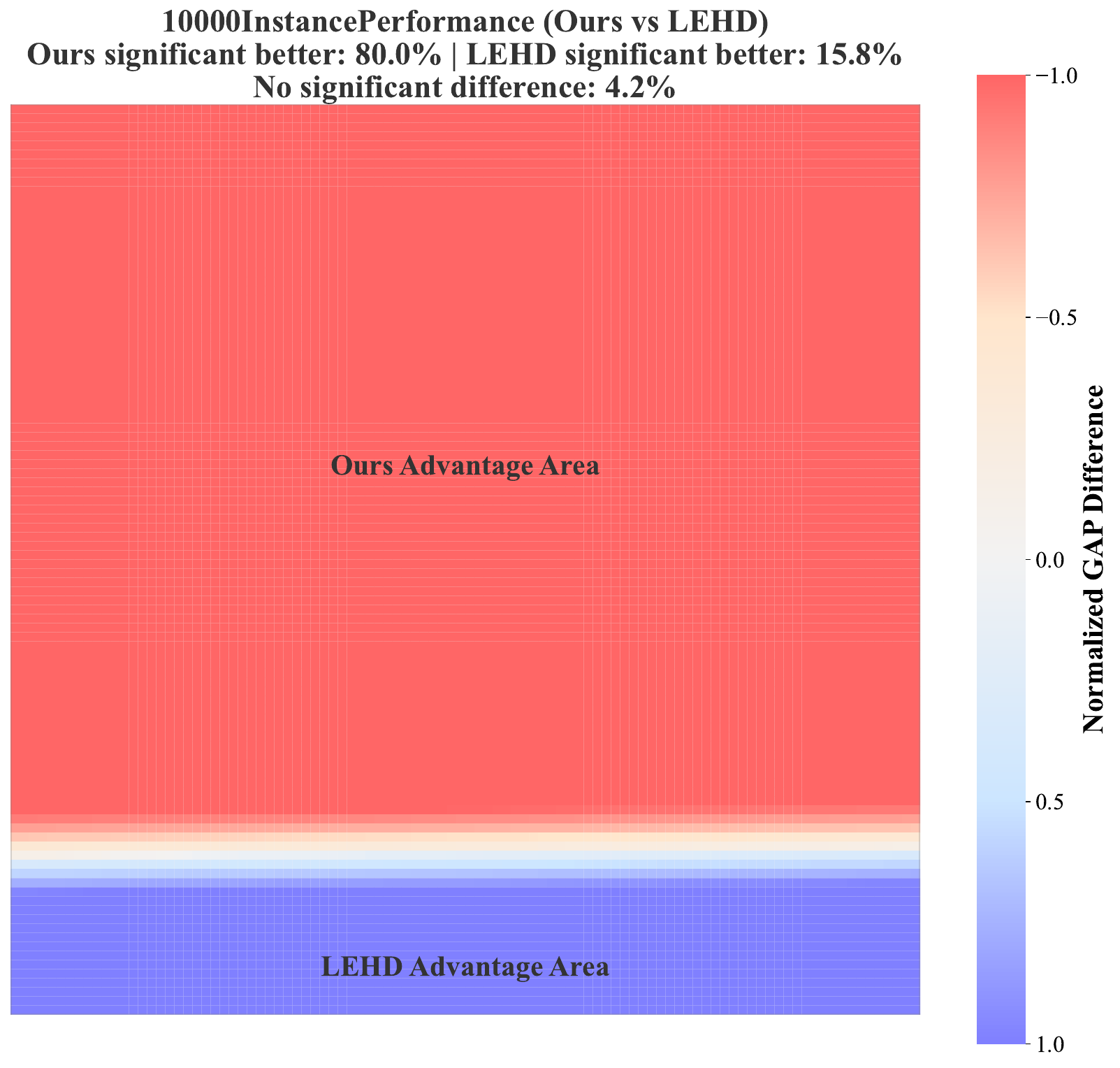}
    \caption{Performance gaps over 10,000 XML100 instances. A 0.5\% gap defines significance: red denotes our model outperforming LEHD, and blue denotes POMO outperforming ours.}
    \label{fig:heatmap}
\end{figure}

In our experiments, we further assessed the performance of the heavy decoder baseline model LEHD at the instance level. The comparative heatmap, ordered by gap percentage, is depicted in Fig.~\ref{fig:heatmap}. Our proposed model surpassed the baseline heavy decoder SOTA LEHD in 82.2\% of the 10,000 instances featuring complex distributions, with 80.0\% exhibiting a significant advantage (i.e., a gap difference exceeding the 0.5\% significance threshold). These findings underscore that, although stacking attention layers within the decoder offers advantages for large-scale tasks, it performs poorly on complex OOD problems. 

Examining the instance distribution, we analyzed the top 100 instances where our model exhibits a clear advantage. These cases typically feature uniform or small node demands, clustered customer nodes, extremely long decision paths, and depot locations that are either central or random. The performance gaps in our favor consistently exceed 20\%, with the maximum reaching 126\%. Conversely, we examined the top 100 instances where POMO demonstrates a pronounced lead. These advantages manifest in scenarios with centrally located depots, randomly distributed customer nodes with high demands, and long decision paths. The gaps in POMO's favor range from approximately 20\%, peaking at 119\%.

In summary, across the comprehensive fair XML100 OOD dataset, our model exhibits marginally superior performance and solution stability. However, part of the gap percentage over 10\% indicates that constructive heuristic neural models remain far from optimal under zero- or few-shot generalization.

\subsubsection{Ablation Study}
This subsection reports two ablations for the pretrained graph diffusion model: (i) removing the constraint‑matrix mask from the autoregressive model, and (ii) comparing discrete Bernoulli‑noise diffusion with continuous Gaussian‑noise diffusion for the diffusion model.

\begin{figure}[!ht]
    \centering
    \includegraphics[width=1.0\linewidth]{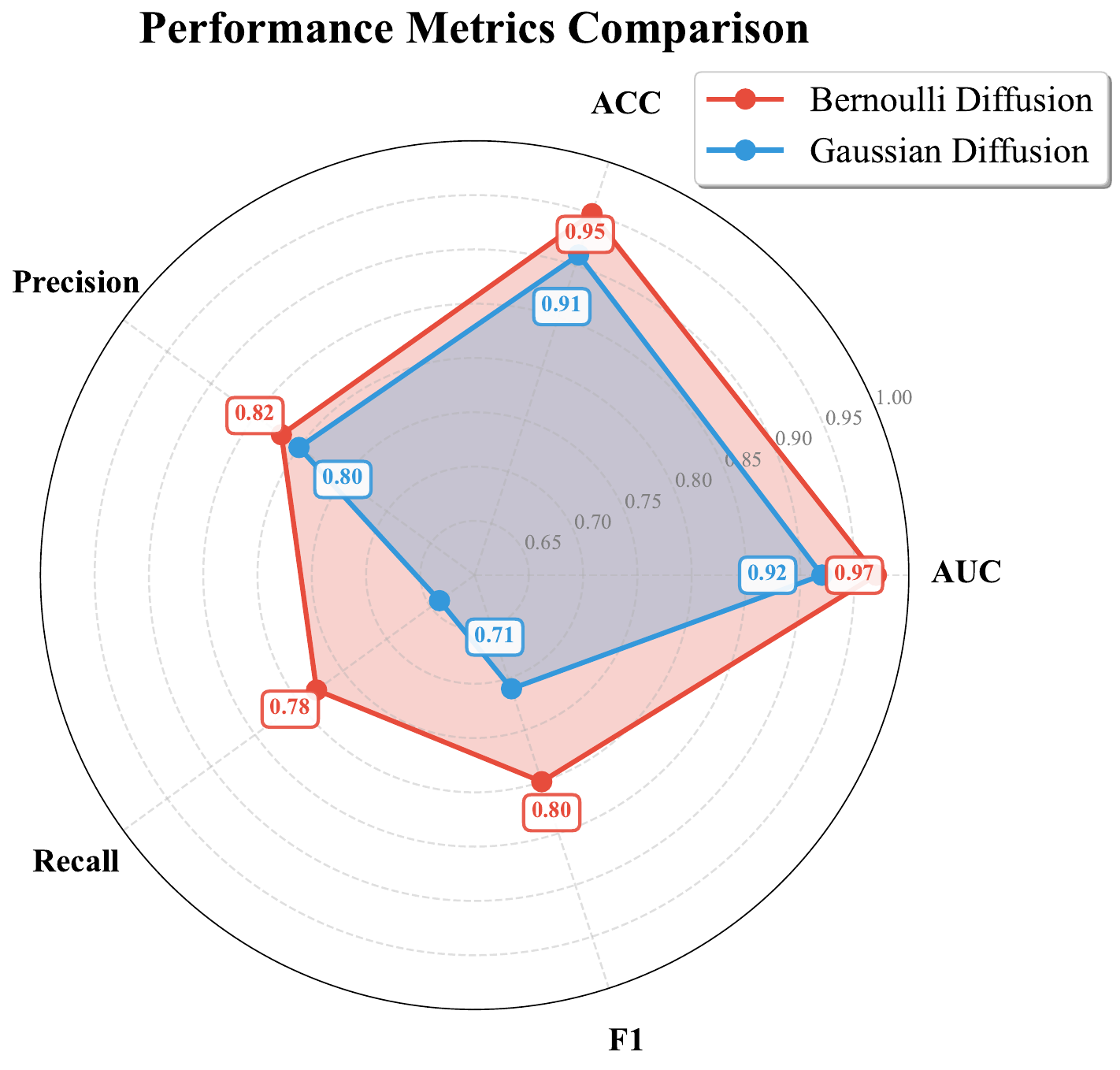}
    \caption{Evaluation metrics for diffusion models parameterized by Bernoulli and Gaussian noise. All metrics are computed from entrywise classification statistics of the constraint matrix.}
    \label{fig:difablation}
\end{figure}

Fig.~\ref{fig:difablation} compares diffusion models with continuous (Gaussian) and discrete (Bernoulli) noise under identical settings. As each constraint‑matrix entry is a binary indicator (constrained vs. unconstrained), we assess mask discrimination on the validation sets using standard classification metrics. With graph‑network hyperparameters fixed, the discrete model better suits constraint‑matrix generation. This accords with the instance graph’s discrete, sparse adjacency: under high entropy, discrete‑state Markov transitions provide a more appropriate inductive bias than continuous noise.

\begin{figure}[!ht]
    \centering
    \includegraphics[width=1.0\linewidth]{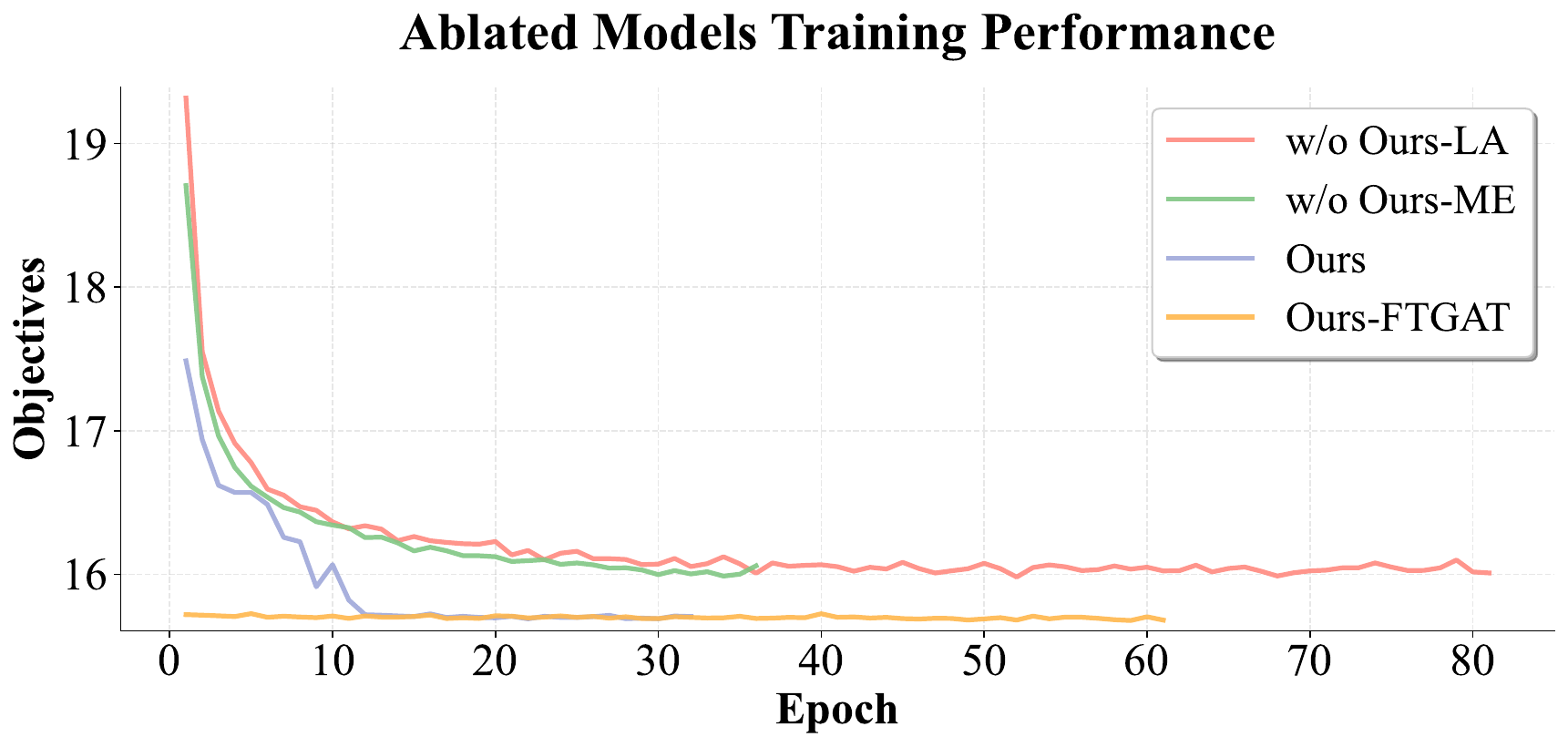}
    \caption{Training curves for the ablation study.}
    \label{fig:ablationstudy}
\end{figure}

We also ablate two components of the autoregressive model: the encoder‑masking module (“Ours‑ME”) and the local‑attention decoder (“Ours‑LA”). We further assess a variant that fine‑tunes the pretrained global GAT during reinforcement learning (“Ours‑FTGAT”). Ablated variants are prefixed “w/o”. Results are reported in Table~\ref{tab:ablationstudy}, and training curves are shown in Fig.~\ref{fig:ablationstudy}. Since early stopping is used, the number of training epochs varies across models.
\begin{table}[tbh]
\caption{Experimental results for the ablation and Fine-Tuned models}
    \label{tab:ablationstudy}
    \centering
    \begin{tabular}{c|cc|c}
    \Xhline{1pt}
      \multirow{2}{*}{Module}   & \multicolumn{2}{c|}{CVRP100} & CVRPLib \\
      \multirow{2}{*}{}& Objectives & Gap(\%)& Average Gap(\%)  \\
      \hline
       Ours  &   15.70 & 0.00 & 5.10\\
       Ours-FTGAT & 15.68 & -0.13 & 5.52 \\
       w/o Ours-ME & 15.92 & 1.40 & 6.33 \\
       w/o Ours-LA & 15.88 & 1.14 & 5.98 \\
       \Xhline{1pt}
    \end{tabular}
    
\end{table}

Ablation studies confirm that applying graph‑diffusion prior masks to both the encoder and decoder yields clear gains for the autoregressive model. We also fine‑tuned the pretrained encoder GAT, achieving additional improvements on synthetic data, nearly matching LKH3 on in‑distribution tests. However, OOD generalization deteriorated, likely due to overfitting from over‑tuning. Accordingly, in the final model we freeze the pretrained GAT to enhance robustness.
\subsubsection{Sensitivity analysis}
\begin{figure}[!ht]
    \centering
    \includegraphics[width=1.0\linewidth]{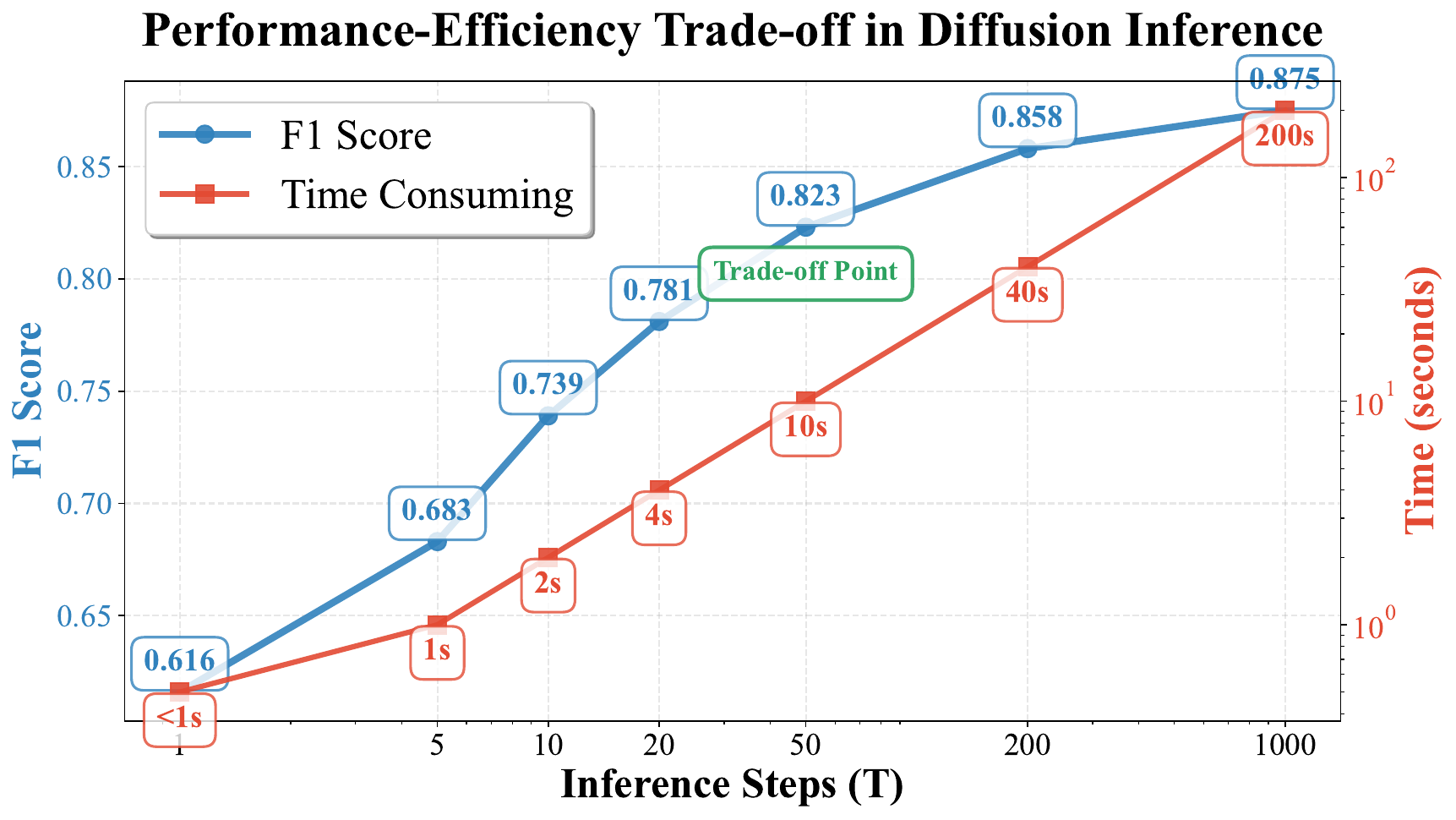}
    \caption{Sensitivity analysis of inference steps in the diffusion model
}
    \label{fig:difstep}
\end{figure}

In our hyperparameter analysis, we vary two factors: the number of diffusion inference steps \(T'\) and the number of graph neural network layers \(L\). For simplicity, all GATs are assigned the same depth \(L\), as their role across modules is to extract node representation embeddings. Although the diffusion model is trained with a linear noise schedule of \(T=1000\), at inference we employ step-skipping to recover denoised outputs with a reduced \(T'\). Unlike DDIM~\cite{song2020denoising} and Sun et al~\cite{sun2023difusco}, we do not perform per-step sampling, as it is prohibitively time-consuming; our goal is for the pre-trained diffusion model to rapidly provide the prior constraint mask required by the autoregressive model. A grid search over \(T'\) (Fig.~\ref{fig:difstep}) indicates that 50 steps offer a favorable accuracy efficiency trade-off. Since the autoregressive component dominates inference time, choosing \(T' > 50\) is not economical for online deployment.

\begin{figure}[!ht]
    \centering
    \includegraphics[width=1.0\linewidth]{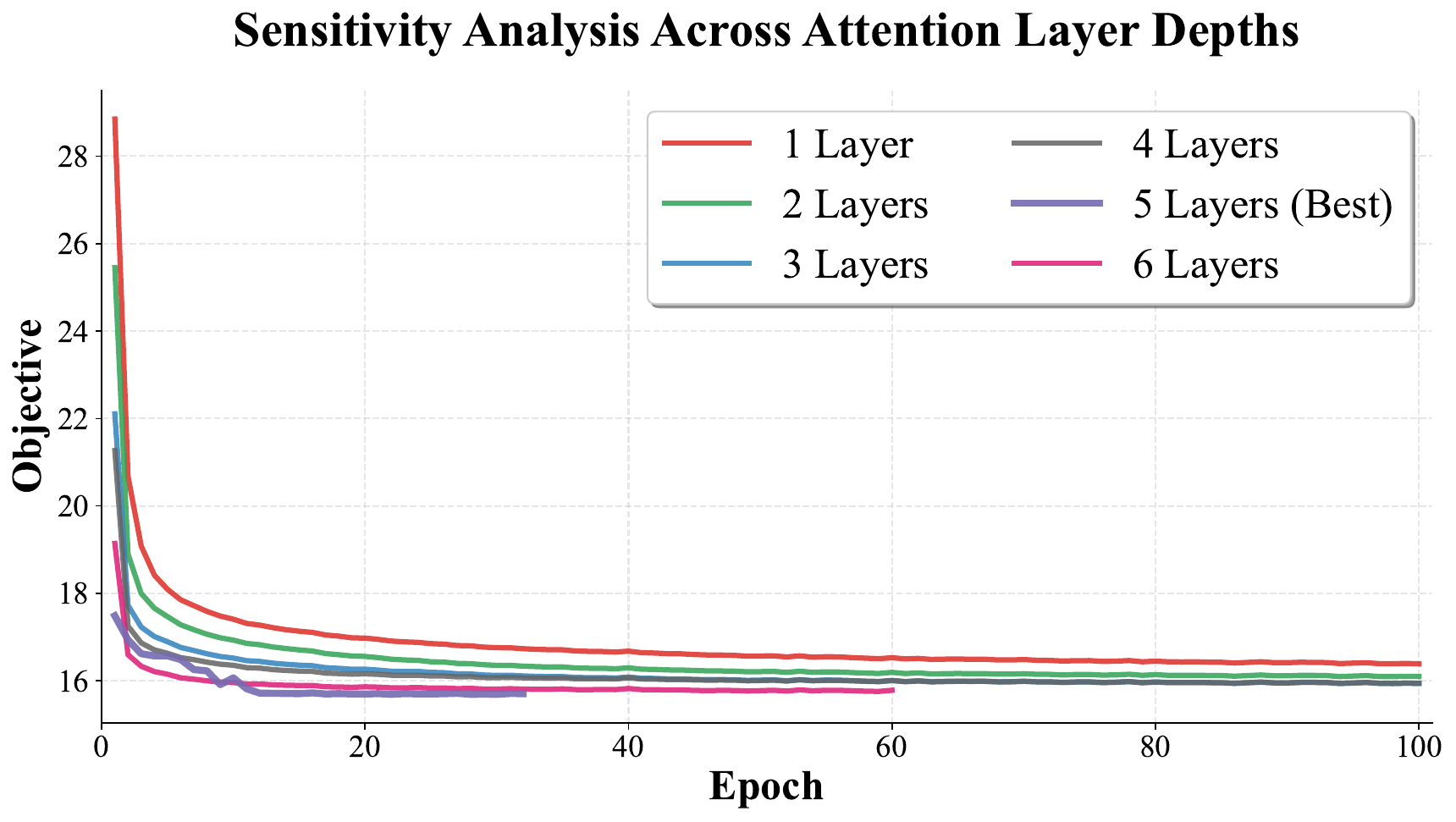}
    \caption{Training curves of the encoder GAT across depths: 1–4‑layer models did not meet the early‑stopping criterion within 100 epochs, whereas 5–6‑layer models triggered early stopping.}
    \label{fig:attenlayers}
\end{figure}

Fig.~\ref{fig:attenlayers} examines the effect of GAT depth (number of layers) on model performance. Moderate stacking effectively captures multi-scale feature dependencies; however, excessive depth introduces spurious long-range aggregation and noise during representation learning. For instance, features of nodes separated by large spatial distances may be erroneously fused by deep attention stacks, leading to diffuse attention at decision time. Moreover, overly deep GATs exacerbate representation oversmoothing, particularly under clustered customer layouts or uniformly distributed demands.

\subsubsection{Discussion and Future Work}
We summarize findings for these three research questions. For RQ1, on synthetic datasets matched to training (20, 50, 100 nodes), our model outperforms eight autoregressive baselines (ten variants). Without inference augmentation, it sets a new SOTA with a 2.52\% gain; augmentation adds 0.19\% at the cost of slower inference.

For RQ2, on CVRPLIB, our method achieves average SOTA for instances under 200 nodes (0.35\% gain), but performance declines with increasing size due to limitations of diffusion‑generated constraints and insufficient decoder depth; current re‑decoding architectures are better suited to large scales. On XML100 (a first careful zero‑shot evaluation), our model surpasses existing autoregressive solvers and markedly mitigates oversmoothing from spatial/demand clustering, with gains often over 20\% in combined‑distribution settings. Nonetheless, very long route‑length decisions remain challenging: even at 100 nodes, OOD gaps can exceed 10\%.

For RQ3, ablations validate the effectiveness of the graph‑diffusion prior (constraint matrix) and its encoder–decoder fusion. Performance is not strongly sensitive to the examined hyperparameters, and the chosen settings are supported by enumeration/grid search.

Recent advances emphasize large‑scale, cross‑distribution optimization and variant ensembles. Yet on small/medium scales, fast‑inference neural solvers still show limited zero-/few‑shot generalization and do not consistently dominate under matched distributions. We advocate robust methods for small‑ to medium‑scale settings across heterogeneous feature distributions and Mixture of Experts (MOE) neural solvers tailored to instance‑level structure. We hope these insights catalyze further progress in neural combinatorial optimization.
\section{Conclusion}
We propose a discrete-noise graph diffusion model to generate a constraint-matrix prior for the vehicle routing problem. This prior is integrated with an autoregressive encoder–decoder via a unified global–local masked attention mechanism. The framework achieves state-of-the-art performance on synthetic datasets and public benchmarks. Comprehensive analyses on datasets with complex feature distributions show that our method markedly mitigates oversmoothing of node embeddings, particularly under tight customer clustering and limited demand heterogeneity, while maintaining strong generalization. Our framework exposes the performance ceilings of existing autoregressive models. Despite the benefits of supervised constrained priors, they fail to guarantee instance-level optimality in CVRP tasks. Thus, incorporating MOE mechanisms represents a promising direction for future work.

\section*{Acknowledgments}


\bibliographystyle{IEEEtran}
\bibliography{LaTeX/aaai2026}

\newpage


\vfill

\end{document}